\documentclass{article}

\usepackage[preprint]{neurips_2023}
\usepackage[utf8]{inputenc} %
\usepackage[T1]{fontenc}    %
\usepackage{hyperref}       %
\usepackage{url}            %
\usepackage{booktabs}       %
\usepackage{amsfonts}       %
\usepackage{amsmath}        % needed for \binom
\usepackage{nicefrac}       %
\usepackage{microtype}      %
\usepackage{xcolor}         %
\usepackage{graphicx}
\usepackage{float}          % retained for fixed table placement
\usepackage{placeins} % keep figures with the relevant major section
\newcommand{\scititle}{LLM Multi-Agent Systems Are Vulnerable to Herding and Runaway Cascades}

\title{\scititle}

\author{%
Farnoosh Hashemi$^{1}$ \qquad Shiyu Ji$^{1}$ \qquad Joice Chen$^{2}$ \qquad Michael W. Macy$^{1}$ \\
$^1$Cornell University \quad $^2$Stanford University
}
\begin{document}

\maketitle

\begin{abstract}
Large language models (LLMs) increasingly operate in multi-agent systems where they collaborate by responding to others' choices, creating endogenous feedback loops with vulnerability to social influence, herding behavior and runaway cascades. Following the design of the classic Music Lab experiment, we tested this vulnerability in large populations of LLM agents that recommended unfamiliar songs, with or without exposure to cumulative recommendation counts. Exposure led to runaway cascades that generated extreme and unpredictable collective recommendations increasingly divergent from a song's intrinsic appeal. Compared to humans, LLM populations were generally far more susceptible to herding behavior, and unlike humans, LLM cascades were less able to self-correct. This vulnerability has implications for multi-agent collaboration in recommender systems, automated evaluation, complex problem-solving, and unauthorized conspiracies.
\end{abstract}

\section{Introduction}

Large language models (LLMs) increasingly operate in multi-agent systems in
which each model's decision becomes an input for other models, creating
endogenous feedback loops with self-reinforcing collective behavior
\citep{rahwan2019machine, grossmann2023ai, zheng2023judging,
park2023generative, brinkmann2023machine}. Yet little is known about the
vulnerability of these systems to runaway cascades that fail to self-correct.
Evaluating an isolated model may reveal whether it produces accurate, useful,
or biased judgments on its own but not how the collective behavior of agent
populations shapes the subsequent individual choices that then aggregate as
collective behavior.

In human populations, theories of herd behavior and information cascades show
how people may discount their own judgment after observing the choices of
others \citep{banerjee1992simple, bikhchandani1992theory}. An initial advantage
can make an option more visible or credible, attracting further choices that
strengthen the original signal. This cumulative advantage can concentrate
success among a few alternatives \citep{merton1968matthew}, leading to
unpredictable outcomes that are highly sensitive to early choices
\citep{salganik2006experimental}.

In the classic ``Music Lab'' study, Salganik, Dodds, and Watts placed
participants in parallel online markets containing the same 48 previously
unfamiliar songs. In each of two experiments, eight ``social influence worlds''
displayed accumulated download counts, whereas one ``independent world''
omitted them. With exposure to social influence, initial popularity became
self-reinforcing, turning an arbitrary early leader into a ``cultural winner,''
even with relatively weak intrinsic appeal in the independent condition.
Compared to the independent world, success was both more unequal and more
unpredictable in the social influence worlds, especially when songs were listed
in order of popularity rather than arranged randomly on a grid
\citep{salganik2006experimental}.

Nevertheless, songs with high intrinsic appeal were able to eventually recover
\citep{salganik2008leading}, and a reanalysis of the Music Lab data found that
popularity became increasingly aligned with intrinsic appeal as choices
accumulated, indicating that the cascade could self-correct
\citep{vanderijt2019self}. Related experiments found that social influence
reduced the diversity underlying the ``wisdom of crowds''
\citep{lorenz2011social}, that one arbitrary positive rating generated herding
\citep{muchnik2013social}, that small arbitrary initial advantages produced
further success across multiple social domains \citep{vanderijt2014field}, and
that variability in early choices led partisan populations to switch sides on
policy issues from one trial to another \citep{macy2019opinion}. As with Music
Lab, subsequent analyses of some of these studies revealed a tendency for
errant cascades to self-correct.

The same feedback structure increasingly appears in human interaction with AI
agents and algorithmic systems. Individual LLMs are known to favor responses
that agree with a user's stated position even when agreement conflicts with
accuracy, a behavior described as sycophancy \citep{sharma2024towards}, and
feedback between humans and AI systems could amplify initially small AI biases
beyond the amplification observed among humans \citep{glickman2025human}. In
recommender systems, past user behavior shapes what algorithms make visible,
and exposure reinforces subsequent behavior. These feedback loops have been
shown to alter the diversity of consumption \citep{fleder2009blockbuster},
homogenize choices, reduce utility \citep{chaney2018algorithmic}, and amplify
existing popularity biases \citep{mansoury2020feedback}.

A growing literature examines whether LLMs can approximate human collective
judgments, reproduce aggregate survey patterns, and replicate findings from
human behavioral experiments \citep{argyle2023out, aher2023using}. However,
LLMs remain imperfect substitutes for human data
\citep{santurkar2023whose, bisbee2024synthetic}, and similarity in aggregate
responses does not establish that LLMs reproduce human social dynamics.
Interacting LLM agents have been found to exhibit consensus and fragmentation
dynamics \citep{chuang2024simulating}, to conform to peer responses to a degree
that varied with interaction structure \citep{han2026conformity}, and to
spontaneously develop shared conventions, collective biases, and tipping
dynamics in which committed minorities shifted population-wide norms
\citep{flintashery2025emergent}.

Experiments on conformity and sycophancy have tested LLM susceptibility to
social influence using a majority answer, peer judgment, or user opinion as an
externally supplied input \citep{sharma2024towards, han2026conformity}.
However, social influence does not necessarily lead to a cascade without a
feedback loop created by an endogenous social signal, and a cascade may fail to
self-correct, depending on the strength of social influence relative to
independent judgment.

The risk of herding behavior and runaway cascades in multi-agent AI systems
motivates seven research questions:

\begin{enumerate}
  \item Does exposure to collective behavior generate self-reinforcing herd behavior?
  \item Does herding generate errant cascades that diverge from independent judgments?
  \item Are cascades sensitive to initial choices that make outcomes unpredictable?
  \item Does susceptibility to cascade failure increase with the strength of social influence?
  \item Do these cascades eventually self-correct?
  \item Are LLMs more susceptible to cascades compared to a human benchmark?
  \item Are reasoning models and models with higher intelligence less susceptible compared to other models?
\end{enumerate}

In our study, the social signal is generated within the
population: recommendations by earlier agents become the popularity counts
observed by later agents. Using replicate influence worlds and an independent
baseline, we test whether LLMs are susceptible to cascades that fail to
self-correct, measure changes in this susceptibility as we vary the strength
of social influence, and compare LLM collective behavior with a human
benchmark. As multi-agent AI systems continue to rapidly proliferate, these
contributions will be increasingly relevant. A more detailed discussion of related work is provided in appendix~\ref{app:related}.

\section{Methods}

\begin{figure}[H]
    \centering
    \includegraphics[width=\textwidth]{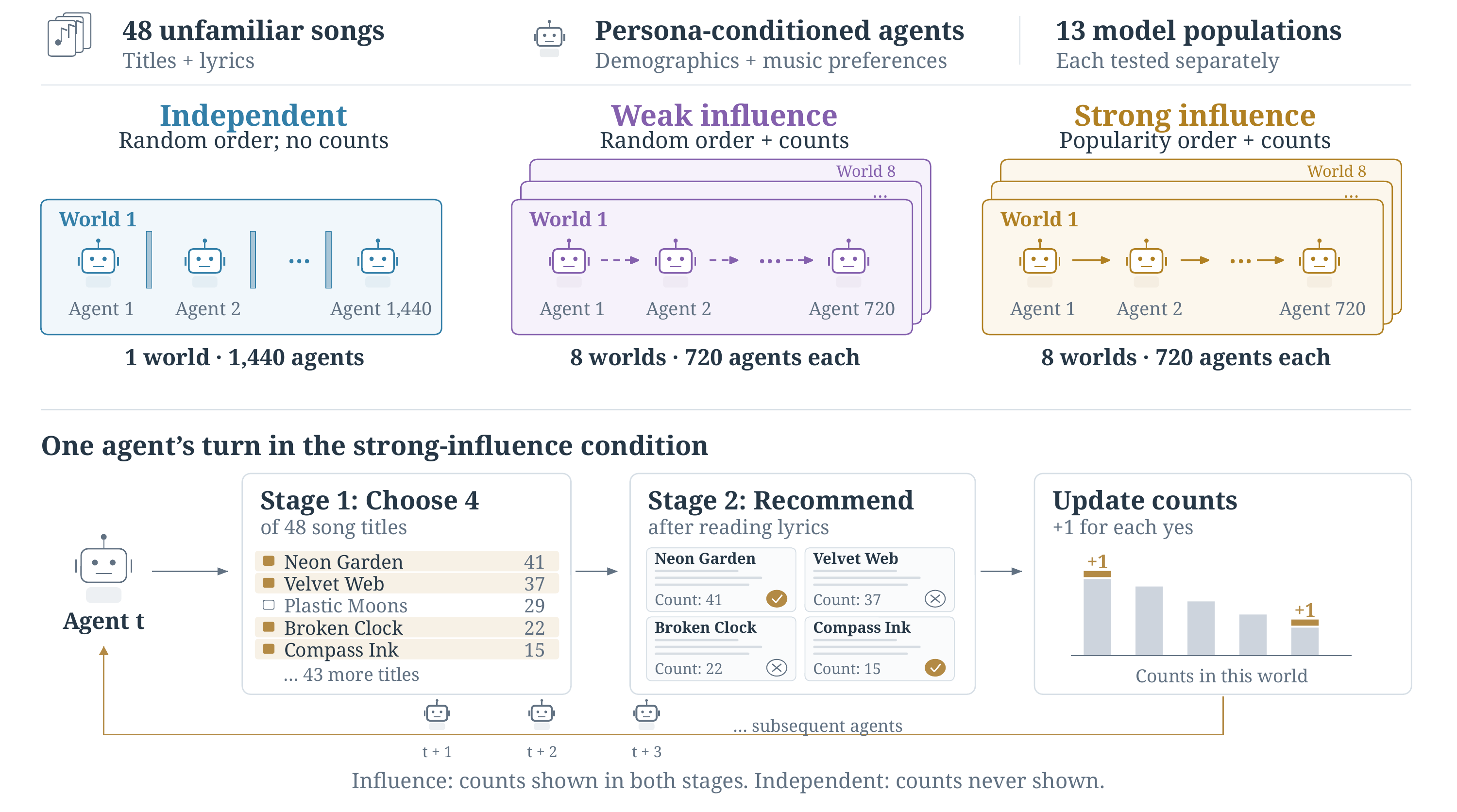}
\caption{\textbf{Experimental design.} Each of 13 model populations evaluated
the same 48 unfamiliar songs, with agents assigned personas combining
demographic attributes and preferred music genres. For each model, one independent world contained 1,440 agents,
and eight worlds in each influence condition contained 720 agents each. Titles
were randomly ordered without recommendation counts in the independent
condition, randomly ordered with counts under weak influence, and ordered by
decreasing counts under strong influence. Walls indicate the absence of social
feedback; dashed and solid arrows indicate feedback under weak and strong
influence, respectively. The bottom sequence illustrates one agent's turn in
the strong-influence condition. Agents first selected four titles, then read
the lyrics and decided whether to recommend each song. In the influence
conditions, cumulative counts were visible in both the selection and
recommendation stages and never shown in the independent condition. Each
recommendation increased the song's count by one, updating the information
available to subsequent agents in the same world. All worlds began with zero
counts and evolved separately.}
\label{fig:design}
\end{figure}

We studied whether feedback from earlier LLM recommendations changes
subsequent choices and collective outcomes. Agents sequentially selected and
evaluated items from a shared catalog; in the influence conditions, their
recommendations updated counts observed by later agents. The social signal was
therefore endogenous, capturing a feedback structure in which model-generated
evaluations become inputs to subsequent decisions. Following the parallel-world
design of Music Lab \citep{salganik2006experimental}, independent decisions
provided a model-specific baseline, while replicate influence worlds allowed us
to measure variation in collective outcomes under the same model, stimuli, and
procedures (fig.~\ref{fig:design}).

\subsection{Task and stimuli}

All models evaluated the same 48 songs, newly generated using Suno v4.5
\citep{suno}. Each stimulus consisted of a title and lyrics, without audio. Newly generated
songs reduced the risk that training exposure or established popularity would
introduce an external signal into the experiment. Lyrics provided a common
text input across all models and made large-scale evaluation practical,
aligning the task with language-based recommendation and automated evaluation.

Each persona-conditioned agent completed two stages, issued as separate
prompts (figs.~\ref{fig:prompt_influence} and \ref{fig:prompt_independent}). In the selection stage, it viewed all 48 titles and selected
exactly four distinct songs to assess. In the recommendation stage,
it received the full lyrics of those songs and made a yes/no recommendation
decision for each, allowing zero to four recommendations. In the influence
conditions, current recommendation counts were visible at both stages. After
the agent completed its decisions, each affirmative recommendation increased
that song's count by one before the next agent acted. Unselected songs could
not receive a recommendation from that agent.

The fixed selection budget standardized evaluation opportunities across
agents; four approximated the average number of songs sampled by Music Lab
participants \citep{salganik2006experimental}. The budget also captured
a bottleneck relevant to LLM applications: downstream evaluation can only act
on candidates selected upstream. Recommender systems separate candidate
generation from ranking \citep{covington2016deep}; retrieval-augmented
generation (RAG) conditions on a restricted set of retrieved documents
\citep{lewis2020retrieval}; and LLM rankers evaluate retrieved candidates
\citep{hou2024large}. Our task captures this shared selection constraint.
Popularity can therefore affect both which items enter the consideration set
and how those items are evaluated, allowing these influences to accumulate
through sequential feedback. 

\subsection{Parallel worlds and feedback conditions}

For each model, we created 17 worlds: one independent world of 1,440 agents,
eight weak-influence worlds of 720 agents, and eight strong-influence worlds
of 720 agents. This yielded 12,960 agents per model and 168,480 agents across
13 models. All worlds began with zero recommendation counts and evolved
separately. In the independent world, agents received no information about
the choices of others. In the influence worlds, agents saw the number of recommendations
each song had accumulated from previous agents in the same world, updated
after every agent. Agents acted sequentially, were unaware of other worlds,
and did not exchange messages. Earlier choices were conveyed only through
cumulative counts and, under strong influence, the resulting title order.

In the independent condition, titles were randomly ordered for each
agent and recommendation counts were omitted. In the weak-influence
condition, counts were displayed, but title order was randomized independently
for each agent. In the strong-influence condition, counts were
displayed and titles were ordered by decreasing count, with ties broken
randomly. The strong-influence prompt also stated that titles were ordered by
recommendation counts. Comparing weak influence with independent evaluation
isolated the effect of displaying counts; comparing strong with weak influence
tested the additional effect of linking popularity to list position. Because LLM
outputs can depend on where information appears in the input
\citep{liu2024lost} and LLM rankers exhibit position and popularity biases
\citep{hou2024large}, this contrast tested whether making popular items more prominent in
the input further strengthened feedback. The same independent world served as
the baseline for both influence conditions. Replicate worlds shared the same
model, stimuli, persona-generating procedure, and inference settings, but each
accumulated its own recommendation history. Comparing their variability with
the independent baseline tested whether feedback amplified variation beyond
that observed under independent evaluation.

\subsection{Model populations}

We evaluated five reasoning models (GPT-5, GPT-5 nano, Gemini 3 Flash,
DeepSeek-R1, and Claude Sonnet 4.6) and eight non-reasoning models (GPT-4o,
GPT-4o mini, Gemma 3 4B, Gemma 3 12B, Gemma 3 27B, DeepSeek-V3.1,
Claude Haiku 4.5, and Qwen3-Next 80B), classified according to their
developers' designations of the inference mode. Each population used a single
model. We also assessed capability using the Artificial Analysis Intelligence
Index v4.1.1 \citep{artificialanalysis}, a composite benchmark spanning
agentic tasks, coding, general problem-solving, and scientific reasoning
(table~\ref{tab:models}).

To introduce structured heterogeneity, each agent was assigned a persona
combining demographic attributes and primary and secondary preferred music
genres. The same persona-generating procedure was used across models and
conditions. Persona construction, complete prompts,
and inference settings are provided in appendix~\ref{app:methods}.

\subsection{Outcome measures}\label{sec:measures}

A song's market share was the proportion of all recommendations it received
within its world:
\begin{equation}
p_{w}(i) = r_{i}^{(w)}/R^{(w)},
\end{equation}
where $r_i^{(w)}$ is the number received by song $i$ in world $w$ and
$R^{(w)}=\sum_{i=1}^{K}r_i^{(w)}$ is the total over all $K=48$ songs.
Market shares in the independent world provided a population-specific measure
of each song's relative intrinsic appeal. 

Our primary outcome measures were Kullback--Leibler (KL) divergence and
unpredictability. $D_{\mathrm{KL}}$ measures how far the distribution of
recommendations in each influence world departed from the independent
distribution:
\begin{equation}
D_{\mathrm{KL}}^{(w)} = \sum_{i=1}^{K}\widetilde{p}_{w}(i)
\log\frac{\widetilde{p}_{w}(i)}{\widetilde{p}_{\mathrm{ind}}(i)},
\end{equation}
where $\widetilde{p}$ denotes market shares after Jeffreys smoothing
(pseudocount 0.5), applied to both distributions (appendix~\ref{app:measures}).
Larger values indicate greater divergence from independent
evaluation. We report the arithmetic mean across the eight worlds in each
influence condition.

Unpredictability ($U$) measures the extent to which parallel worlds with the
same model, songs, and experimental procedures evolved toward different
collective outcomes \citep{salganik2006experimental}:
\begin{equation}
u_{i} = \binom{W}{2}^{-1}\sum_{j=1}^{W-1}\sum_{k=j+1}^{W}
\left| p_{j}(i) - p_{k}(i) \right|,
\qquad
U = \frac{1}{K}\sum_{i=1}^{K}u_{i},
\end{equation}
where $W=8$ and market shares were unsmoothed. For the independent world, we
randomly divided agents into two groups of 720, matching the size of an
influence world, and averaged their market-share differences over 1,000
splits.

To test for self-correction, we followed the procedure applied to the Music Lab
data \citep{vanderijt2019self}. For each influence world, we split the
chronological recommendation sequence into two approximately equal halves and
computed the Spearman correlation between song recommendation counts within
each half and independent-world counts. Second-half counts excluded
first-half recommendations. A positive change,
$\Delta\rho^{(w)}=\rho_2^{(w)}-\rho_1^{(w)}>0$, indicates that later
recommendations aligned more closely with intrinsic appeal.

Together, these measures distinguish displacement from independent judgment,
variation across replicate worlds, and self-correction over time. For the
capability analysis, we used relative unpredictability,
$U_{\mathrm{inf}}/U_{\mathrm{ind}}$, to measure proportional amplification
above each model's independent baseline; absolute $U$ remained the primary
measure in the other analyses.
Additional measures (Gini inequality, Pearson correlation with independent shares, and persistence of early recommendations), robustness checks, and position-bias analyses are reported in the
Supplementary Materials.

\subsection{Human benchmark}\label{sec:human}
 
We applied the same outcome definitions to the original Music Lab data as a
human benchmark \citep{salganik2006experimental}, with downloads in place of
recommendations. Weak influence corresponded to Music Lab experiment 1, in which
download counts were shown but songs were arranged in a randomized grid, and
strong influence corresponded to experiment 2, in which songs were listed by
download count. Because stimuli, response
tasks, and presentation differed, the human results provide a qualitative
benchmark rather than a controlled estimate of human--LLM differences;
appendix~\ref{app:human} details this comparison.
\FloatBarrier
\section{Results}

\subsection{LLM susceptibility to herding}

\begin{figure}[H]
    \centering
    \includegraphics[width=\textwidth]{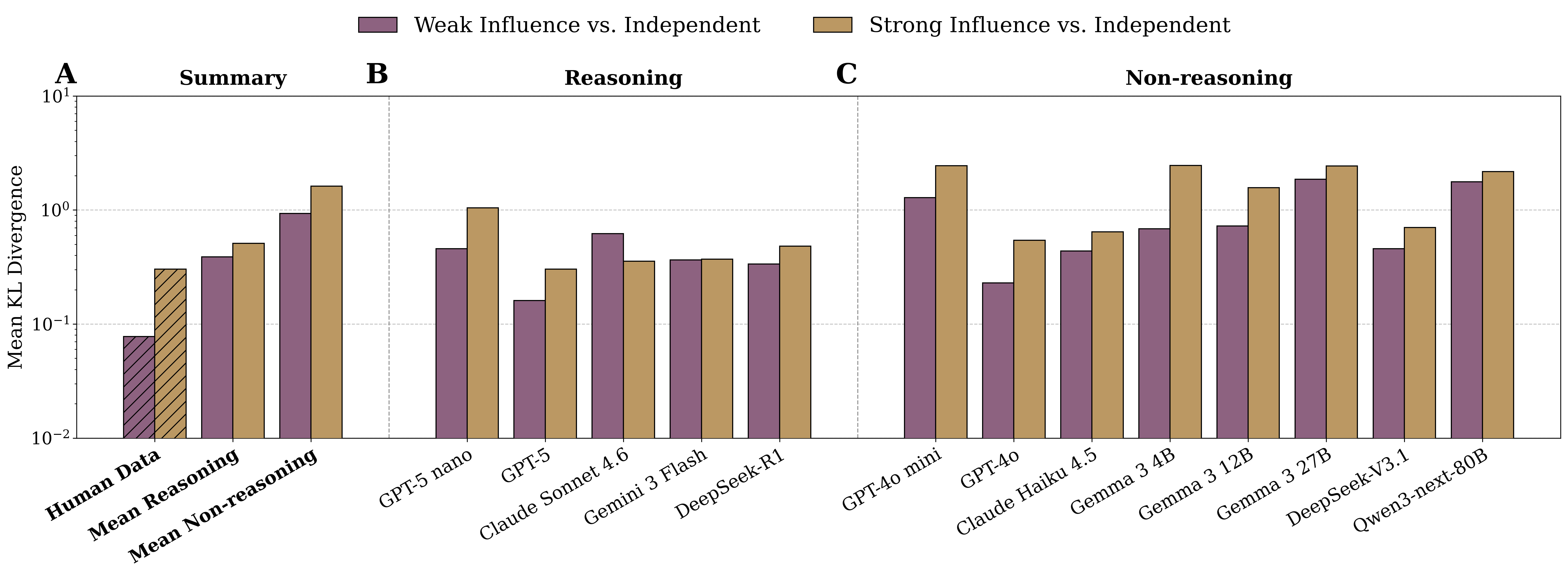}
\caption{\textbf{Compared with humans, LLM populations showed greater divergence
from independent judgment.} Kullback--Leibler (KL) divergence measures
departure of influence-world market shares from the corresponding independent
baseline, averaged over eight worlds per condition. Higher values on the
logarithmic vertical axis indicate greater divergence. Divergence was computed
with Jeffreys smoothing (pseudocount 0.5 per song). Gold and purple denote
strong and weak influence, respectively. Hatched bars show human values from
the original Music Lab data \citep{salganik2006experimental}. \textbf{(A)} Mean
divergence was greater among non-reasoning models than among reasoning models,
and greater in both classes than among humans. \textbf{(B} and \textbf{C)}.
GPT-5 nearly matched the human benchmark under strong influence; Gemma 3 4B had
the greatest divergence under strong influence and Gemma 3 27B under weak
influence. Differences between independent and social influence conditions
were significant for all 13 LLMs under both influence conditions
($p<0.001$).}
\label{fig:kl_divergence}
\end{figure}

Fig.~\ref{fig:kl_divergence} shows how exposure to current recommendations
generated herding behavior in which agents' recommendations diverged from
independent judgment ($p<0.001$ for every model under both influence
conditions). Mean KL divergence was about nine times the human benchmark under
weak influence and four times under strong influence
(fig.~\ref{fig:kl_divergence}A). Among LLMs, non-reasoning models diverged 2.4
and 3.2 times as far as reasoning models under weak and strong influence,
respectively. Under strong influence, four of the five reasoning models
diverged less than every non-reasoning model; GPT-5 nano was the exception
(fig.~\ref{fig:kl_divergence}, B and C). GPT-5 approached human divergence,
whereas Gemma 3 4B diverged more than eight times as far. Except for Claude
Sonnet 4.6, strong influence increased mean divergence by 65\%, with greater
divergence in 12 of 13 models (two-sided Wilcoxon signed-rank test across
models, $p<0.01$).

\subsection{Social feedback can turn herding behavior into a stampede}

\begin{figure}[H]
    \centering
    \includegraphics[width=\textwidth]{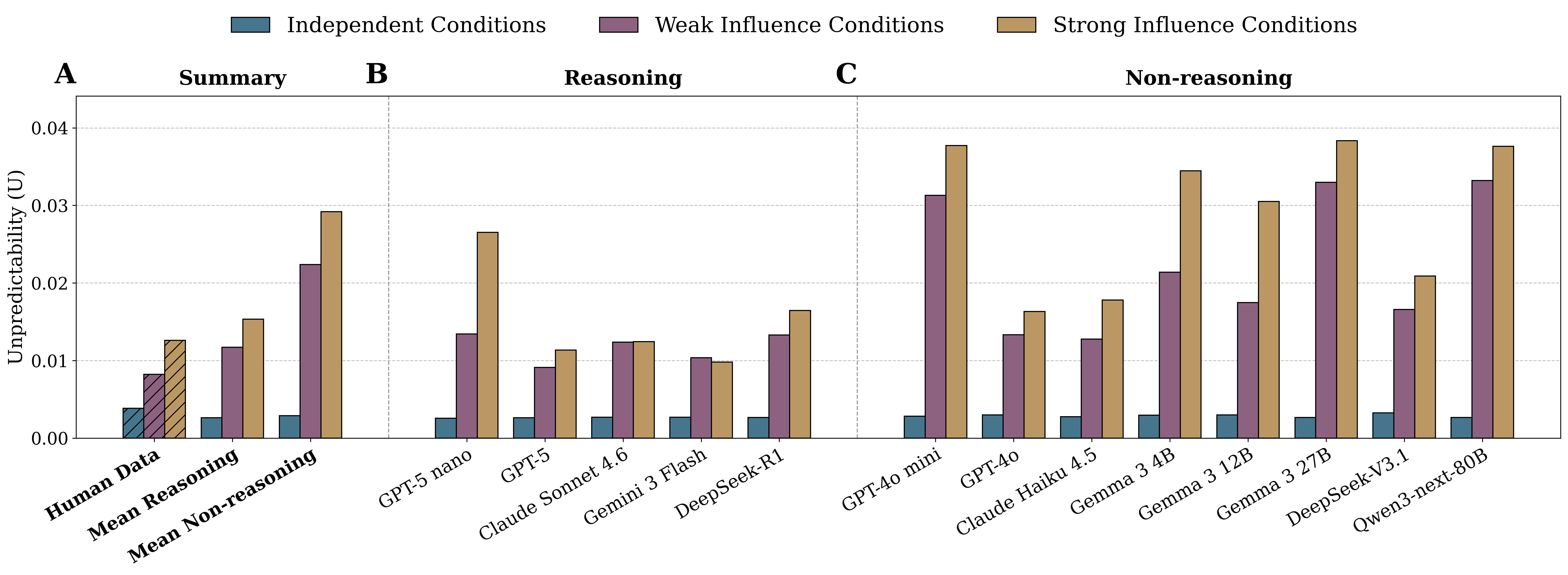}
\caption{\textbf{Social influence increased unpredictability, especially among
non-reasoning models.} As in the original Music Lab experiment
\citep{salganik2006experimental}, unpredictability (U) measures the mean
absolute difference in a song's market share across all 28 pairs of eight
influence worlds, averaged over the 48 songs. Unpredictability in the
independent world was estimated by randomly splitting the population into two
equal groups of 720 agents and averaging over 1,000 random splits. Gold,
purple, and blue denote strong influence, weak influence, and independent
evaluation, respectively. The hatched bar is the human benchmark for comparison
with unweighted class means among the five reasoning and eight non-reasoning
models. \textbf{(A)} Unpredictability was greater among non-reasoning models
compared to reasoning models and humans, regardless of influence level.
\textbf{(B} and \textbf{C)} Model-specific results show that unpredictability
in the independent condition was near zero for each model and only slightly
higher among humans, indicating stable aggregate shares under independnet evaluation.
Social influence increased unpredictability in every model, although some
reasoning models fell slightly below the human benchmark. In every population,
unpredictability was greater in both influence conditions than in the
independent condition ($p<0.001$), except for humans ($p<0.01$) and
DeepSeek-R1 under weak influence ($p<0.05$).}
\label{fig:unpredictability}
\end{figure}

Divergence from independent judgment can indicate herding behavior but not
necessarily a stampede. The latter also requires an endogenous feedback loop
that can amplify small initial differences, making outcomes highly
unpredictable. Fig.~\ref{fig:unpredictability}A shows that unpredictability in
the independent condition was close to zero across all models and slightly
below the level observed among humans. Compared to the independent condition,
unpredictability was far greater in the influence conditions for humans and
LLMs, especially among non-reasoning models in the strong-influence condition.

Given the similarly low independent baselines, the more instructive comparison
is within influence conditions. Non-reasoning models were nearly twice as
unpredictable as reasoning models at both influence levels. Mean
unpredictability among reasoning models was modestly above the human
benchmark, whereas non-reasoning models were more than twice as unpredictable
as humans under both influence conditions.

All 13 models exceeded human unpredictability under weak influence
(fig.~\ref{fig:unpredictability}, B and C). Under strong influence, three
reasoning models---Gemini 3 Flash, GPT-5, and Claude Sonnet 4.6---fell slightly
below the human benchmark. Strong influence increased unpredictability in 12 of
13 models, with Gemini 3 Flash the lone exception (two-sided Wilcoxon
signed-rank test across models, $p<0.001$).

Unpredictability reflects sensitivity to early choices, which can also be
measured directly.  We  also correlated market shares among the first
30 agents in each influence world with market shares among all later agents
in the same world. In LLM populations, early movers predicted later popularity
better than intrinsic appeal did in 12 of 13 models under weak influence and in
all 13 under strong influence, with the largest gaps among non-reasoning models
(table~\ref{tab:pearson_results}). Among humans, early movers were the better
predictor only under strong influence (appendix~\ref{app:early_movers}).

\subsection{Susceptibility to herding and cascades decreased with model intelligence scores.}

\begin{figure}[H]
    \centering
    \includegraphics[width=0.9\textwidth]{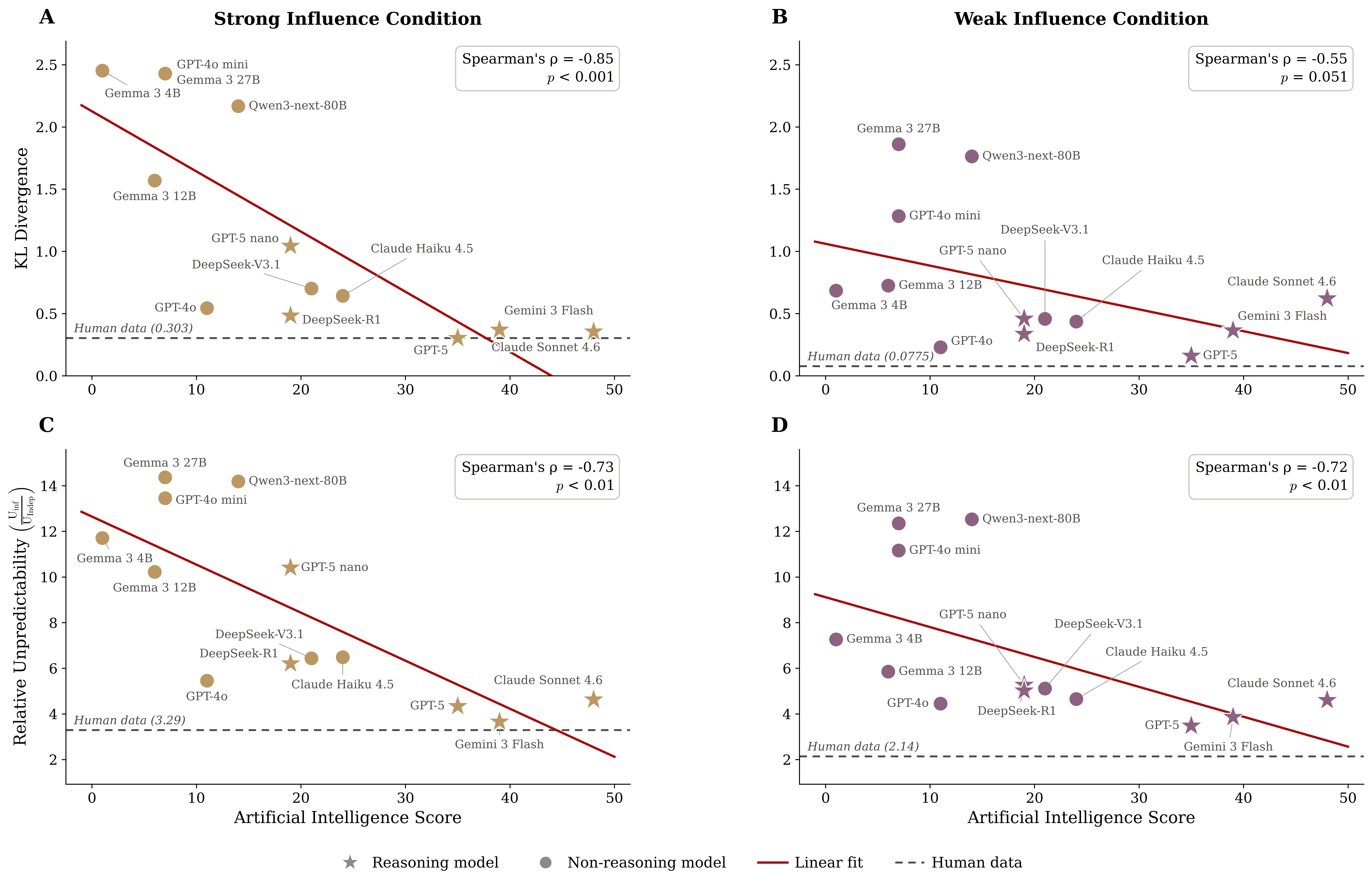}
\caption{\textbf{Higher model intelligence is associated with lower divergence
and relative unpredictability.} Each point represents one of 13 LLMs plotted
against its Artificial Analysis Intelligence Index score, version 4.1.1
\citep{artificialanalysis}. The index aggregates performance across benchmark
agentic tasks, coding, general problem-solving, and scientific reasoning;
higher scores indicate higher intelligence. The associations are summarized as
a least-squares linear fit and Spearman correlation. Stars denote reasoning
models and dots denote non-reasoning. The dashed line is the human benchmark.
Within each influence condition, KL divergence is the mean over eight influence
worlds and unpredictability is the mean over 28 pairs of influence worlds,
normalized by unpredictability in the independent condition
($U_{\mathrm{inf}}/U_{\mathrm{ind}}$). The association with divergence
depends on the strength of social influence, going from $\rho=-0.851$
($p<0.001$) when influence is strong \textbf{(A)} to $\rho=-0.551$
($p=0.051$) when influence is weak \textbf{(B)}. The association with
unpredictability \textbf{(C} and \textbf{D)} remains strong regardless of
influence condition ($\rho=-0.733$, $p<0.01$ and $\rho=-0.719$,
$p<0.01$ for strong and weak influence, respectively). The models are
clustered by reasoning class, indicating the expected association with
intelligence score.}
\label{fig:capability_u}
\end{figure}

A continuous measure of model intelligence provides a more refined analysis
compared to the binary reasoning classification. Fig.~\ref{fig:capability_u}
reports results using the Artificial Analysis Intelligence Index (AAII), an
established composite performance benchmark. Stars and dots denote reasoning
and non-reasoning models, respectively. The dashed line is the human benchmark.
Each association is summarized with a least squares linear fit and Spearman
correlation. Across the 13 models, both KL divergence
(fig.~\ref{fig:capability_u}A and B) and relative unpredictability
(fig.~\ref{fig:capability_u}C and D) decline as intelligence scores increase,
regardless of influence condition (although the association is stronger for
divergence in the strong-influence condition). For KL divergence,
$\rho=-0.851$ and $\rho=-0.551$ for strong and weak influence,
respectively. For relative unpredictability, $\rho=-0.733$ and
$\rho=-0.719$. Models are clustered by reasoning class, as expected.

\subsection{Human cascades self-corrected; LLM cascades generally did not}

\begin{figure}[H]
    \centering
    \includegraphics[width=0.89\textwidth]{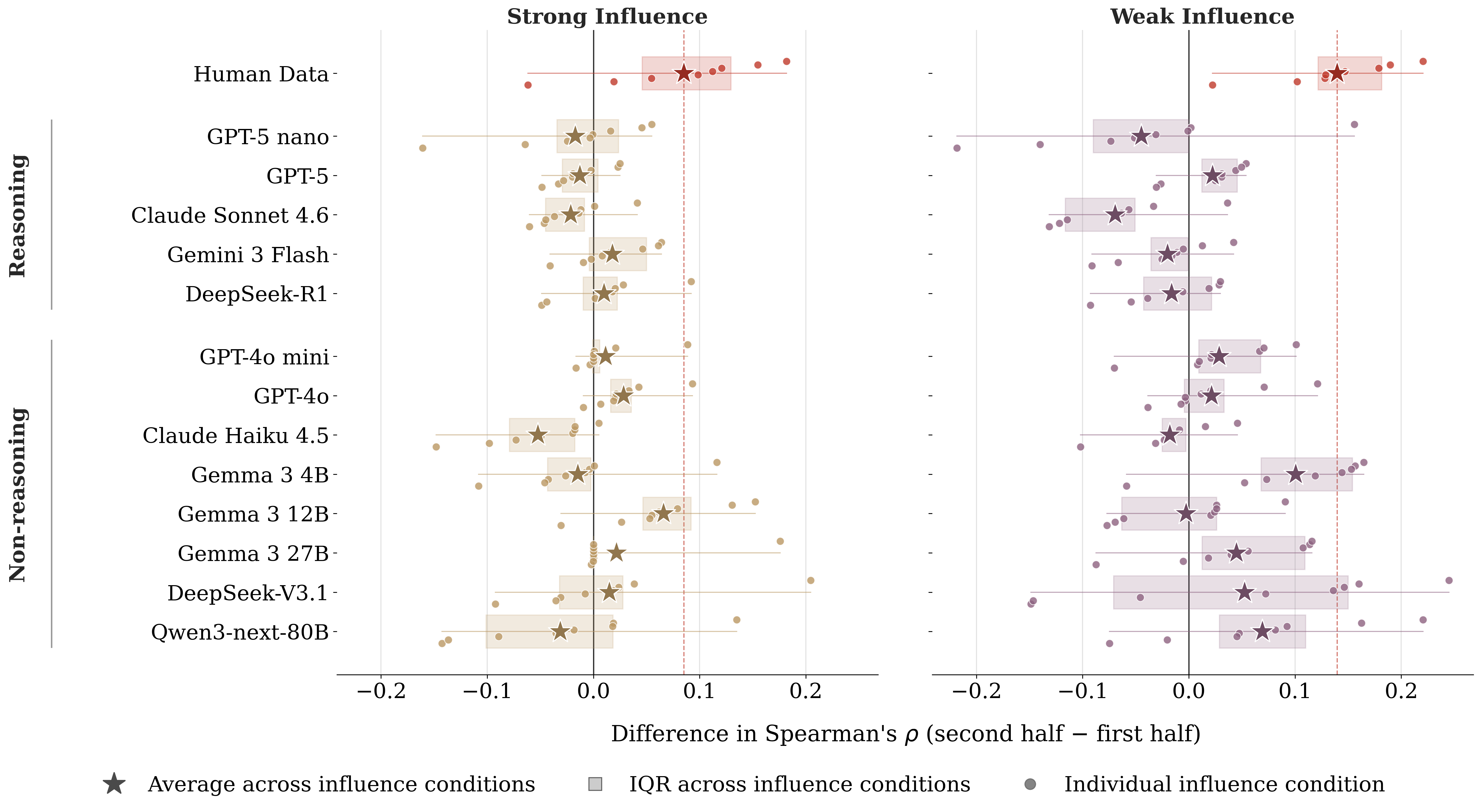}
\caption{\textbf{Human cascades consistently self-corrected; LLM cascades
generally did not.} Following van de Rijt \citep{vanderijt2019self},
self-correction was measured as the change in Spearman correlation between
popularity and intrinsic appeal between the first and second halves of the
sequential recommendations. The change in correlation was calculated for each
of the eight influence worlds in each influence condition (indicated by a dot)
and then averaged over the eight worlds (indicated by a star). A higher
second-half correlation indicates closer alignment with the independent
ranking; negative changes indicate movement of popularity away from intrinsic
appeal. Shaded boxes and lines show the interquartile range and total range
respectively. Human correlations increased in seven of eight strong-influence
worlds and all eight weak-influence worlds, with mean changes of +0.085
($p<0.05$) and +0.140 ($p<0.01$), respectively. In contrast, LLM changes
varied in direction across models and worlds and averaged near zero under
strong influence. As expected, self-correction requires intrinsic appeal that
is strong relative to social influence.}
\label{fig:self_correction}
\end{figure}

Cascades can self-correct if independent preferences are sufficiently strong
relative to social influence \citep{vanderijt2019self}.
Fig.~\ref{fig:self_correction} reports self-correction using the same
diagnostic procedure that was used with the original Music Lab data -- the
change in the Spearman correlation between popularity and intrinsic appeal
between the first and second halves of the experiment. A larger second-half
correlation indicates movement toward the independent ranking; negative
changes indicate movement away.

Across the entire experiment, market shares remained positively correlated with
intrinsic appeal in every population, but this correspondence weakened in 11 of
13 models under strong influence (table~\ref{tab:pearson_results}). Splitting
each experiment in half, human cascades self-corrected in 15 of 16 influence
worlds, with the popularity-appeal correlation increasing from $\rho=0.633$
to 0.773 under weak influence (all eight worlds; one-sided Wilcoxon signed-rank
test, $p<0.01$) and from 0.575 to 0.660 under strong influence (seven of
eight worlds; $p<0.05$). Among LLMs, changes varied in direction across
models and worlds, averaging close to zero under strong influence. The human
benchmark therefore exhibited a corrective dynamic that was much less
consistent across LLM populations. In short, LLMs were more susceptible to
``runaway cascades,'' persistent, history-dependent departures without reliable
self-correction over the observed experimental horizon. However, the results do
not establish that correction could never occur over longer sequences.

\subsection{Supplementary results}
Social influence increased inequality (measured as the Gini coefficient) in all
13 models under strong influence and in 12 under weak influence
(fig.~\ref{fig:gini}). Relative to their independent baselines, non-reasoning
models showed larger mean increases in inequality than humans, whereas
reasoning models showed smaller increases.
 
Popularity-ordered lists in the strong-influence condition could confound
responses to recommendation counts with responses to display position. We
tested for position bias using the fraction of first-stage selections at each
of the 48 display positions (not songs) in the independent condition
(fig.~\ref{fig:position_bias}). Selections were approximately uniform across
positions for most LLMs, although Gemma 3 4B and GPT-4o mini showed a modest
preference for higher-ranked positions. Because agents were randomly
distributed across conditions, those in the influence conditions should not be
expected to have greater position bias than those in the independent
condition. As expected, position bias was much stronger among humans, given the
lack of the cognitive resources to process all 48 songs. Net of position bias,
susceptibility to self-reinforcing social influence among humans may be
substantially weaker than we reported, which means we may have understated the
difference in susceptibility with LLMs. See ``Position bias'' in the SM for
details and an extended discussion of the implications.

Chain-of-thought reasoning traces in DeepSeek-R1 revealed explicit references
to song popularity in every log and in both influence conditions, consistent
with responsiveness to social influence.

Finally, we ran two robustness checks. Under strong influence, lowering the
temperature from 1.0 to 0.7 did not attenuate overall divergence or
unpredictability among non-reasoning models
(fig.~\ref{fig:temperature_robustness}). Increasing the number of titles that
could be chosen for assessment from four to 12 reduced divergence and
unpredictability but preserved the main patterns across models
(figs.~\ref{fig:selection_kl_robustness} and
\ref{fig:selection_u_robustness}).

\FloatBarrier
\section{Discussion}

Large language models increasingly collaborate in multi-agent systems in which
model-generated judgments may become inputs for subsequent agents, creating
endogenous feedback loops. Evaluating collective AI behavior therefore requires
going beyond individual responses to a fixed signal to test how model outputs
shape subsequent inputs.

Across 13 experiments, comparisons with a human benchmark revealed generally
greater LLM susceptibility to herding and cascades and less consistent
self-correction. When popularity influences recommendation, songs with an
early advantage are more likely to be chosen for assessment and to then be
assessed more favorably.

Reasoning and higher-intelligence models were less susceptible to herding
behavior, but lower susceptibility did not ensure self-correction. Going
forward, as models rapidly advance, susceptibility to influence on simple tasks
like recommendation can be expected to decline. However, our experiment
precluded more powerful influence mechanisms like communication, task
incentives, and social sanctioning. If task complexity increases alongside
intelligence and influence is reinforced with social pressure and incentives,
susceptibility to errant cascades may increase.

Our design preserved the feedback structure of Music Lab but adapted its
implementation to LLM agents: agents read the lyrics of newly generated songs
rather than listening to songs by unknown bands, recommended rather than
downloaded songs, assessed a fixed four songs rather than choosing how many to
sample, and saw a text list rather than a grid when songs were not ordered by
popularity (appendix~\ref{app:human}). These differences limit attribution of
human--LLM contrasts to population type. A matched human replication of this study is needed to rule out these design differences as the explanation for observed behavioral
differences. Our experiment also studied homogeneous populations of a single
model, precluded explicit collaboration, and lacked incentives for conformity
or ground-truth accuracy. Generalization to heterogeneous teams, more
challenging tasks, multimodal systems, incentives for compliance, and verbal
influence remain to be tested. Nevertheless, self-reinforcing departures from
independent judgment identify a potential risk for recommender systems,
automated evaluation, and collaborative problem-solving.

\section*{Acknowledgments}

This study was supported by the U.S. National Science Foundation
(SBE 2611776) and Amazon Research during the time this study was conducted.

\FloatBarrier
\bibliography{bib.bib}

\begin{thebibliography}{67}
\providecommand{\natexlab}[1]{#1}
\providecommand{\url}[1]{\texttt{#1}}
\expandafter\ifx\csname urlstyle\endcsname\relax
  \providecommand{\doi}[1]{doi: #1}\else
  \providecommand{\doi}{doi: \begingroup \urlstyle{rm}\Url}\fi

\bibitem[Aher et~al.(2023)Aher, Arriaga, and Kalai]{aher2023using}
Gati~V. Aher, Rosa~I. Arriaga, and Adam~Tauman Kalai.
\newblock Using large language models to simulate multiple humans and replicate human subject studies.
\newblock In \emph{Proceedings of the 40th International Conference on Machine Learning}, volume 202 of \emph{Proceedings of Machine Learning Research}, pages 337--371, 2023.

\bibitem[Argyle et~al.(2023)Argyle, Busby, Fulda, Gubler, Rytting, and Wingate]{argyle2023out}
Lisa~P. Argyle, Ethan~C. Busby, Nancy Fulda, Joshua~R. Gubler, Christopher Rytting, and David Wingate.
\newblock Out of one, many: Using language models to simulate human samples.
\newblock \emph{Political Analysis}, 31\penalty0 (3):\penalty0 337--351, 2023.
\newblock \doi{10.1017/pan.2023.2}.

\bibitem[{Artificial Analysis}(2026)]{artificialanalysis}
{Artificial Analysis}.
\newblock Artificial analysis intelligence index v4.1.
\newblock \url{https://artificialanalysis.ai/articles/artificial-analysis-intelligence-index-v4-1}, 2026.

\bibitem[Ashokkumar et~al.(2026)Ashokkumar, Hewitt, Ghezae, and Willer]{ashokkumar2026large}
Ashwini Ashokkumar, Luke Hewitt, Isaias Ghezae, and Robb Willer.
\newblock Large language models can predict the results of social science experiments.
\newblock \emph{Nature}, 656\penalty0 (8126):\penalty0 115--122, 2026.
\newblock \doi{10.1038/s41586-026-10742-x}.

\bibitem[Baltaji et~al.(2024)Baltaji, Hemmatian, and Varshney]{baltaji2024conformity}
Razan Baltaji, Babak Hemmatian, and Lav Varshney.
\newblock Conformity, confabulation, and impersonation: Persona inconstancy in multi-agent {LLM} collaboration.
\newblock In Vinodkumar Prabhakaran, Sunipa Dev, Luciana Benotti, Daniel Hershcovich, Laura Cabello, Yong Cao, Ife Adebara, and Li~Zhou, editors, \emph{Proceedings of the 2nd Workshop on Cross-Cultural Considerations in {NLP}}, pages 17--31, Bangkok, Thailand, August 2024. Association for Computational Linguistics.
\newblock \doi{10.18653/v1/2024.c3nlp-1.2}.
\newblock URL \url{https://aclanthology.org/2024.c3nlp-1.2/}.

\bibitem[Banerjee(1992)]{banerjee1992simple}
Abhijit~V Banerjee.
\newblock A simple model of herd behavior.
\newblock \emph{The quarterly journal of economics}, 107\penalty0 (3):\penalty0 797--817, 1992.

\bibitem[Bellina et~al.(2026)Bellina, De~Marzo, and Garcia]{bellina2026conformity}
Alessandro Bellina, Giordano De~Marzo, and David Garcia.
\newblock Conformity and social impact on ai agents.
\newblock \emph{arXiv preprint arXiv:2601.05384}, 2026.

\bibitem[Bikhchandani et~al.(1992)Bikhchandani, Hirshleifer, and Welch]{bikhchandani1992theory}
Sushil Bikhchandani, David Hirshleifer, and Ivo Welch.
\newblock A theory of fads, fashion, custom, and cultural change as informational cascades.
\newblock \emph{Journal of Political Economy}, 100\penalty0 (5):\penalty0 992--1026, 1992.
\newblock \doi{10.1086/261849}.

\bibitem[Bisbee et~al.(2024)Bisbee, Clinton, Dorff, Kenkel, and Larson]{bisbee2024synthetic}
James Bisbee, Joshua~D. Clinton, Cassy Dorff, Brenton Kenkel, and Jennifer~M. Larson.
\newblock Synthetic replacements for human survey data? {The} perils of large language models.
\newblock \emph{Political Analysis}, 32\penalty0 (4):\penalty0 401--416, 2024.
\newblock \doi{10.1017/pan.2024.5}.

\bibitem[Brinkmann et~al.(2023)Brinkmann, Baumann, Bonnefon, Derex, M{\"u}ller, Nussberger, Czaplicka, Acerbi, Griffiths, Henrich, Leibo, McElreath, Oudeyer, Stray, and Rahwan]{brinkmann2023machine}
Levin Brinkmann, Fabian Baumann, Jean-Fran\c{c}ois Bonnefon, Maxime Derex, Thomas~F. M{\"u}ller, Anne-Marie Nussberger, Agnieszka Czaplicka, Alberto Acerbi, Thomas~L. Griffiths, Joseph Henrich, Joel~Z. Leibo, Richard McElreath, Pierre-Yves Oudeyer, Jonathan Stray, and Iyad Rahwan.
\newblock Machine culture.
\newblock \emph{Nature Human Behaviour}, 7\penalty0 (11):\penalty0 1855--1868, 2023.
\newblock \doi{10.1038/s41562-023-01742-2}.

\bibitem[Chaney et~al.(2018)Chaney, Stewart, and Engelhardt]{chaney2018algorithmic}
Allison J.~B. Chaney, Brandon~M. Stewart, and Barbara~E. Engelhardt.
\newblock How algorithmic confounding in recommendation systems increases homogeneity and decreases utility.
\newblock In \emph{Proceedings of the 12th ACM Conference on Recommender Systems}, pages 224--232, 2018.
\newblock \doi{10.1145/3240323.3240370}.

\bibitem[Cho et~al.(2025)Cho, Guntuku, and Ungar]{cho2025herd}
Young-Min Cho, Sharath~Chandra Guntuku, and Lyle Ungar.
\newblock Herd behavior: Investigating peer influence in llm-based multi-agent systems.
\newblock \emph{arXiv preprint arXiv:2505.21588}, 2025.

\bibitem[Choi et~al.(2025)Choi, Kim, Chae, and Baek]{choi2025empirical}
Min Choi, Keonwoo Kim, Sungwon Chae, and Sangyeop Baek.
\newblock An empirical study of group conformity in multi-agent systems.
\newblock In Wanxiang Che, Joyce Nabende, Ekaterina Shutova, and Mohammad~Taher Pilehvar, editors, \emph{Findings of the {A}ssociation for {C}omputational {L}inguistics: {ACL} 2025}, pages 5123--5139, Vienna, Austria, July 2025. Association for Computational Linguistics.
\newblock ISBN 979-8-89176-256-5.
\newblock \doi{10.18653/v1/2025.findings-acl.265}.
\newblock URL \url{https://aclanthology.org/2025.findings-acl.265/}.

\bibitem[Chuang et~al.(2024)Chuang, Goyal, Harlalka, Suresh, Hawkins, Yang, Shah, Hu, and Rogers]{chuang2024simulating}
Yun-Shiuan Chuang, Agam Goyal, Nikunj Harlalka, Siddharth Suresh, Robert Hawkins, Sijia Yang, Dhavan Shah, Junjie Hu, and Timothy~T. Rogers.
\newblock Simulating opinion dynamics with networks of {LLM}-based agents.
\newblock In \emph{Findings of the Association for Computational Linguistics: NAACL 2024}, pages 3326--3346, 2024.
\newblock \doi{10.18653/v1/2024.findings-naacl.211}.

\bibitem[Covington et~al.(2016)Covington, Adams, and Sargin]{covington2016deep}
Paul Covington, Jay Adams, and Emre Sargin.
\newblock Deep neural networks for {YouTube} recommendations.
\newblock In \emph{Proceedings of the 10th ACM Conference on Recommender Systems}, pages 191--198, New York, NY, USA, 2016. Association for Computing Machinery.
\newblock \doi{10.1145/2959100.2959190}.

\bibitem[Craswell et~al.(2008)Craswell, Zoeter, Taylor, and Ramsey]{craswell2008experimental}
Nick Craswell, Onno Zoeter, Michael Taylor, and Bill Ramsey.
\newblock An experimental comparison of click position-bias models.
\newblock In \emph{Proceedings of the 2008 International Conference on Web Search and Data Mining}, pages 87--94. ACM, 2008.
\newblock \doi{10.1145/1341531.1341545}.

\bibitem[{De Marzo} et~al.(2026){De Marzo}, Bellina, Castellano, Priesemann, and Garcia]{demarzo2026conformity}
Giordano {De Marzo}, Alessandro Bellina, Claudio Castellano, Viola Priesemann, and David Garcia.
\newblock Conformity generates collective misalignment in {AI} agents societies.
\newblock arXiv:2605.10721, 2026.
\newblock URL \url{https://arxiv.org/abs/2605.10721}.

\bibitem[Du et~al.(2024)Du, Li, Torralba, Tenenbaum, and Mordatch]{du2024improving}
Yilun Du, Shuang Li, Antonio Torralba, Joshua~B. Tenenbaum, and Igor Mordatch.
\newblock Improving factuality and reasoning in language models through multiagent debate.
\newblock In \emph{Proceedings of the 41st International Conference on Machine Learning}, ICML'24. JMLR.org, 2024.

\bibitem[Fanous et~al.(2025)Fanous, Goldberg, Agarwal, Lin, Zhou, Xu, Bikia, Daneshjou, and Koyejo]{fanous2025syceval}
Aaron Fanous, Jacob Goldberg, Ank Agarwal, Joanna Lin, Anson Zhou, Sonnet Xu, Vasiliki Bikia, Roxana Daneshjou, and Sanmi Koyejo.
\newblock Syceval: Evaluating llm sycophancy.
\newblock In \emph{Proceedings of the AAAI/ACM Conference on AI, Ethics, and Society}, volume~8, pages 893--900, 2025.

\bibitem[Filandrianos et~al.(2025)Filandrianos, Dimitriou, Lymperaiou, Thomas, and Stamou]{filandrianos2025bias}
Giorgos Filandrianos, Angeliki Dimitriou, Maria Lymperaiou, Konstantinos Thomas, and Giorgos Stamou.
\newblock Bias beware: The impact of cognitive biases on llm-driven product recommendations.
\newblock In \emph{Proceedings of the 2025 Conference on Empirical Methods in Natural Language Processing}, pages 22408--22437, 2025.

\bibitem[Fleder and Hosanagar(2009)]{fleder2009blockbuster}
Daniel Fleder and Kartik Hosanagar.
\newblock Blockbuster culture's next rise or fall: The impact of recommender systems on sales diversity.
\newblock \emph{Management Science}, 55\penalty0 (5):\penalty0 697--712, 2009.
\newblock \doi{10.1287/mnsc.1080.0974}.

\bibitem[{Flint Ashery} et~al.(2025){Flint Ashery}, Aiello, and Baronchelli]{flintashery2025emergent}
Ariel {Flint Ashery}, Luca~Maria Aiello, and Andrea Baronchelli.
\newblock Emergent social conventions and collective bias in {LLM} populations.
\newblock \emph{Science Advances}, 11\penalty0 (20):\penalty0 eadu9368, 2025.
\newblock \doi{10.1126/sciadv.adu9368}.

\bibitem[Fr{\"o}hling et~al.(2025)Fr{\"o}hling, Demartini, and Assenmacher]{frohling-etal-2025-personas}
Leon Fr{\"o}hling, Gianluca Demartini, and Dennis Assenmacher.
\newblock Personas with attitudes: Controlling {LLM}s for diverse data annotation.
\newblock In Agostina Calabrese, Christine de~Kock, Debora Nozza, Flor~Miriam Plaza-del Arco, Zeerak Talat, and Francielle Vargas, editors, \emph{Proceedings of the 9th Workshop on Online Abuse and Harms ({WOAH})}, pages 468--481, Vienna, Austria, August 2025. Association for Computational Linguistics.
\newblock ISBN 979-8-89176-105-6.
\newblock URL \url{https://aclanthology.org/2025.woah-1.43/}.

\bibitem[Gaubert et~al.(2026)Gaubert, Devaux, {\c{C}}elen, Marjieh, Mangalagiu, Jardin, and Jacoby]{gaubert2026experimental}
L{\'e}na Gaubert, R{\'e}mi Devaux, Elif {\c{C}}elen, Raja Marjieh, Diana Mangalagiu, Antoine Jardin, and Nori Jacoby.
\newblock An experimental method to study opinion diffusion in human-ai hybrid societies.
\newblock \emph{arXiv preprint arXiv:2605.09197}, 2026.

\bibitem[Glickman and Sharot(2025)]{glickman2025human}
Moshe Glickman and Tali Sharot.
\newblock How human--{AI} feedback loops alter human perceptual, emotional and social judgements.
\newblock \emph{Nature Human Behaviour}, 9\penalty0 (2):\penalty0 345--359, 2025.
\newblock \doi{10.1038/s41562-024-02077-2}.

\bibitem[Grossmann et~al.(2023)Grossmann, Feinberg, Parker, Christakis, Tetlock, and Cunningham]{grossmann2023ai}
Igor Grossmann, Matthew Feinberg, Dawn~C. Parker, Nicholas~A. Christakis, Philip~E. Tetlock, and William~A. Cunningham.
\newblock {AI} and the transformation of social science research.
\newblock \emph{Science}, 380\penalty0 (6650):\penalty0 1108--1109, 2023.
\newblock \doi{10.1126/science.adi1778}.

\bibitem[Han et~al.(2026)Han, Tan, Yu, Zheng, and Tang]{han2026conformity}
C.~Han, J.~Tan, B.~Yu, W.~Zheng, and X.~Tang.
\newblock Conformity dynamics in {LLM} multi-agent systems: The roles of topology and self-social weighting.
\newblock arXiv:2601.05606, 2026.
\newblock URL \url{https://arxiv.org/abs/2601.05606}.

\bibitem[Hashemi and Macy(2025)]{hashemi2025collective}
Farnoosh Hashemi and Michael Macy.
\newblock Collective social behaviors in llms: an analysis of llms social networks.
\newblock In \emph{Large Language Models for Scientific and Societal Advances}, 2025.

\bibitem[Hashemi and Macy(2026)]{hashemi2026empirical}
Farnoosh Hashemi and Michael Macy.
\newblock An empirical study of collective behaviors and social dynamics in large language model agents.
\newblock In Vera Demberg, Kentaro Inui, and Llu{\'i}s Marquez, editors, \emph{Proceedings of the 19th Conference of the {E}uropean Chapter of the {A}ssociation for {C}omputational {L}inguistics (Volume 1: Long Papers)}, pages 7327--7351, Rabat, Morocco, March 2026. Association for Computational Linguistics.
\newblock ISBN 979-8-89176-380-7.
\newblock \doi{10.18653/v1/2026.eacl-long.344}.
\newblock URL \url{https://aclanthology.org/2026.eacl-long.344/}.

\bibitem[Hou et~al.(2024)Hou, Zhang, Lin, Lu, Xie, McAuley, and Zhao]{hou2024large}
Yupeng Hou, Junjie Zhang, Zihan Lin, Hongyu Lu, Ruobing Xie, Julian McAuley, and Wayne~Xin Zhao.
\newblock Large language models are zero-shot rankers for recommender systems.
\newblock In \emph{European conference on information retrieval}, pages 364--381. Springer, 2024.

\bibitem[Hussain and Nielbo(2026)]{hussain2026conformity}
Zafar Hussain and Kristoffer Nielbo.
\newblock Conformity mitigations in large language models lie on a single resistance-receptivity frontier.
\newblock \emph{arXiv preprint arXiv:2608.11247}, 2026.

\bibitem[Jain and Krishnamurthy(2025)]{jain2025interacting}
Adit Jain and Vikram Krishnamurthy.
\newblock Interacting large language model agents. bayesian social learning based interpretable models.
\newblock \emph{IEEE Access}, 13:\penalty0 25465--25504, 2025.
\newblock \doi{10.1109/ACCESS.2025.3538599}.

\bibitem[Joachims et~al.(2005)Joachims, Granka, Pan, Hembrooke, and Gay]{joachims2005accurately}
Thorsten Joachims, Laura Granka, Bing Pan, Helene Hembrooke, and Geri Gay.
\newblock Accurately interpreting clickthrough data as implicit feedback.
\newblock In \emph{Proceedings of the 28th Annual International ACM SIGIR Conference on Research and Development in Information Retrieval}, pages 154--161. ACM, 2005.
\newblock \doi{10.1145/1076034.1076063}.

\bibitem[Lewis et~al.(2020)Lewis, Perez, Piktus, Petroni, Karpukhin, Goyal, K{\"u}ttler, Lewis, Yih, Rockt{\"a}schel, Riedel, and Kiela]{lewis2020retrieval}
Patrick Lewis, Ethan Perez, Aleksandra Piktus, Fabio Petroni, Vladimir Karpukhin, Naman Goyal, Heinrich K{\"u}ttler, Mike Lewis, Wen-tau Yih, Tim Rockt{\"a}schel, Sebastian Riedel, and Douwe Kiela.
\newblock Retrieval-augmented generation for knowledge-intensive {NLP} tasks.
\newblock In \emph{Advances in Neural Information Processing Systems}, volume~33, pages 9459--9474. Curran Associates, Inc., 2020.

\bibitem[Lin et~al.(2025)Lin, Huang, Ku, Hsu, Shieh, and Kang]{lin2025towards}
Hsien-Tsung Lin, Pei-Cing Huang, Chan-Tung Ku, Chan Hsu, Pei-Xuan Shieh, and Yihuang Kang.
\newblock Towards simulating social influence dynamics with llm-based multi-agents.
\newblock In \emph{2025 IEEE International Conference on Information Reuse and Integration and Data Science (IRI)}, pages 307--312. IEEE, 2025.

\bibitem[Liu et~al.(2024)Liu, Lin, Hewitt, Paranjape, Bevilacqua, Petroni, and Liang]{liu2024lost}
Nelson~F. Liu, Kevin Lin, John Hewitt, Ashwin Paranjape, Michele Bevilacqua, Fabio Petroni, and Percy Liang.
\newblock Lost in the middle: How language models use long contexts.
\newblock \emph{Transactions of the Association for Computational Linguistics}, 12:\penalty0 157--173, 2024.
\newblock \doi{10.1162/tacl_a_00638}.

\bibitem[Lorenz et~al.(2011)Lorenz, Rauhut, Schweitzer, and Helbing]{lorenz2011social}
Jan Lorenz, Heiko Rauhut, Frank Schweitzer, and Dirk Helbing.
\newblock How social influence can undermine the wisdom of crowd effect.
\newblock \emph{Proceedings of the National Academy of Sciences}, 108\penalty0 (22):\penalty0 9020--9025, 2011.
\newblock \doi{10.1073/pnas.1008636108}.

\bibitem[Macy et~al.(2019)Macy, Deri, Ruch, and Tong]{macy2019opinion}
Michael Macy, Sebastian Deri, Alexander Ruch, and Natalie Tong.
\newblock Opinion cascades and the unpredictability of partisan polarization.
\newblock \emph{Science Advances}, 5\penalty0 (8):\penalty0 eaax0754, 2019.
\newblock \doi{10.1126/sciadv.aax0754}.

\bibitem[Mansoury et~al.(2020)Mansoury, Abdollahpouri, Pechenizkiy, Mobasher, and Burke]{mansoury2020feedback}
Masoud Mansoury, Himan Abdollahpouri, Mykola Pechenizkiy, Bamshad Mobasher, and Robin Burke.
\newblock Feedback loop and bias amplification in recommender systems.
\newblock In \emph{Proceedings of the 29th ACM International Conference on Information \& Knowledge Management}, CIKM '20, page 2145–2148, New York, NY, USA, 2020. Association for Computing Machinery.
\newblock ISBN 9781450368599.
\newblock \doi{10.1145/3340531.3412152}.
\newblock URL \url{https://doi.org/10.1145/3340531.3412152}.

\bibitem[Merton(1968)]{merton1968matthew}
Robert~K. Merton.
\newblock The {Matthew} effect in science.
\newblock \emph{Science}, 159\penalty0 (3810):\penalty0 56--63, 1968.
\newblock \doi{10.1126/science.159.3810.56}.

\bibitem[Muchnik et~al.(2013)Muchnik, Aral, and Taylor]{muchnik2013social}
Lev Muchnik, Sinan Aral, and Sean~J. Taylor.
\newblock Social influence bias: A randomized experiment.
\newblock \emph{Science}, 341\penalty0 (6146):\penalty0 647--651, 2013.
\newblock \doi{10.1126/science.1240466}.

\bibitem[Park et~al.(2023)Park, O'Brien, Cai, Morris, Liang, and Bernstein]{park2023generative}
Joon~Sung Park, Joseph O'Brien, Carrie~Jun Cai, Meredith~Ringel Morris, Percy Liang, and Michael~S. Bernstein.
\newblock Generative agents: Interactive simulacra of human behavior.
\newblock In \emph{Proceedings of the 36th Annual ACM Symposium on User Interface Software and Technology}, UIST '23, New York, NY, USA, 2023. Association for Computing Machinery.
\newblock ISBN 9798400701320.
\newblock \doi{10.1145/3586183.3606763}.
\newblock URL \url{https://doi.org/10.1145/3586183.3606763}.

\bibitem[Peng et~al.(2026)Peng, Tian, Liang, Deng, and Wu]{peng2026engagement}
Tai-Quan Peng, Yuan Tian, Songsong Liang, Dazhen Deng, and Yingcai Wu.
\newblock Do llm-driven agents exhibit engagement mechanisms? controlled tests of information load, descriptive norms, and popularity cues.
\newblock \emph{arXiv preprint arXiv:2603.20911}, 2026.

\bibitem[Perez et~al.(2023)Perez, Ringer, Lukosiute, Nguyen, Chen, Heiner, Pettit, Olsson, Kundu, Kadavath, Jones, Chen, Mann, Israel, Seethor, McKinnon, Olah, Yan, Amodei, Amodei, Drain, Li, Tran-Johnson, Khundadze, Kernion, Landis, Kerr, Mueller, Hyun, Landau, Ndousse, Goldberg, Lovitt, Lucas, Sellitto, Zhang, Kingsland, Elhage, Joseph, Mercado, DasSarma, Rausch, Larson, McCandlish, Johnston, Kravec, El~Showk, Lanham, Telleen-Lawton, Brown, Henighan, Hume, Bai, Hatfield-Dodds, Clark, Bowman, Askell, Grosse, Hernandez, Ganguli, Hubinger, Schiefer, and Kaplan]{perez2023discovering}
Ethan Perez, Sam Ringer, Kamile Lukosiute, Karina Nguyen, Edwin Chen, Scott Heiner, Craig Pettit, Catherine Olsson, Sandipan Kundu, Saurav Kadavath, Andy Jones, Anna Chen, Benjamin Mann, Brian Israel, Bryan Seethor, Cameron McKinnon, Christopher Olah, Da~Yan, Daniela Amodei, Dario Amodei, Dawn Drain, Dustin Li, Eli Tran-Johnson, Guro Khundadze, Jackson Kernion, James Landis, Jamie Kerr, Jared Mueller, Jeeyoon Hyun, Joshua Landau, Kamal Ndousse, Landon Goldberg, Liane Lovitt, Martin Lucas, Michael Sellitto, Miranda Zhang, Neerav Kingsland, Nelson Elhage, Nicholas Joseph, Noemi Mercado, Nova DasSarma, Oliver Rausch, Robin Larson, Sam McCandlish, Scott Johnston, Shauna Kravec, Sheer El~Showk, Tamera Lanham, Timothy Telleen-Lawton, Tom Brown, Tom Henighan, Tristan Hume, Yuntao Bai, Zac Hatfield-Dodds, Jack Clark, Samuel~R. Bowman, Amanda Askell, Roger Grosse, Danny Hernandez, Deep Ganguli, Evan Hubinger, Nicholas Schiefer, and Jared Kaplan.
\newblock Discovering language model behaviors with model-written evaluations.
\newblock In Anna Rogers, Jordan Boyd-Graber, and Naoaki Okazaki, editors, \emph{Findings of the {A}ssociation for {C}omputational {L}inguistics: {ACL} 2023}, pages 13387--13434, Toronto, Canada, July 2023. Association for Computational Linguistics.
\newblock \doi{10.18653/v1/2023.findings-acl.847}.
\newblock URL \url{https://aclanthology.org/2023.findings-acl.847/}.

\bibitem[Pezeshkpour and Hruschka(2024)]{pezeshkpour2024order}
Pouya Pezeshkpour and Estevam Hruschka.
\newblock Large language models sensitivity to the order of options in multiple-choice questions.
\newblock In \emph{Findings of the Association for Computational Linguistics: NAACL 2024}, pages 2006--2017. Association for Computational Linguistics, 2024.
\newblock \doi{10.18653/v1/2024.findings-naacl.130}.

\bibitem[Qu et~al.(2026)Qu, Fu, and Hu]{qu2026easier}
Jiaming Qu, Lucheng Fu, and Yibo Hu.
\newblock Easier to mislead than to correct: Harmful and beneficial revision in llm conformity.
\newblock \emph{arXiv preprint arXiv:2606.01637}, 2026.

\bibitem[Rahwan et~al.(2019)Rahwan, Cebrian, Obradovich, Bongard, Bonnefon, Breazeal, Crandall, Christakis, Couzin, Jackson, et~al.]{rahwan2019machine}
Iyad Rahwan, Manuel Cebrian, Nick Obradovich, Josh Bongard, Jean-Fran{\c{c}}ois Bonnefon, Cynthia Breazeal, Jacob~W Crandall, Nicholas~A Christakis, Iain~D Couzin, Matthew~O Jackson, et~al.
\newblock Machine behaviour.
\newblock \emph{Nature}, 568\penalty0 (7753):\penalty0 477--486, 2019.

\bibitem[Salganik and Watts(2008)]{salganik2008leading}
Matthew~J Salganik and Duncan~J Watts.
\newblock Leading the herd astray: An experimental study of self-fulfilling prophecies in an artificial cultural market.
\newblock \emph{Social psychology quarterly}, 71\penalty0 (4):\penalty0 338--355, 2008.

\bibitem[Salganik et~al.(2006)Salganik, Dodds, and Watts]{salganik2006experimental}
Matthew~J. Salganik, Peter~Sheridan Dodds, and Duncan~J. Watts.
\newblock Experimental study of inequality and unpredictability in an artificial cultural market.
\newblock \emph{Science}, 311\penalty0 (5762):\penalty0 854--856, 2006.
\newblock \doi{10.1126/science.1121066}.

\bibitem[Santurkar et~al.(2023)Santurkar, Durmus, Ladhak, Lee, Liang, and Hashimoto]{santurkar2023whose}
Shibani Santurkar, Esin Durmus, Faisal Ladhak, Cinoo Lee, Percy Liang, and Tatsunori Hashimoto.
\newblock Whose opinions do language models reflect?
\newblock In \emph{Proceedings of the 40th International Conference on Machine Learning}, volume 202 of \emph{Proceedings of Machine Learning Research}, pages 29971--30004, 2023.

\bibitem[Sharma et~al.(2024)Sharma, Tong, Korbak, Duvenaud, Askell, Bowman, Cheng, Durmus, Hatfield-Dodds, Johnston, Kravec, Maxwell, McCandlish, Ndousse, Rausch, Schiefer, Yan, Zhang, and Perez]{sharma2024towards}
Mrinank Sharma, Meg Tong, Tomasz Korbak, David Duvenaud, Amanda Askell, Samuel~R. Bowman, Newton Cheng, Esin Durmus, Zac Hatfield-Dodds, Scott~R. Johnston, Shauna Kravec, Timothy Maxwell, Sam McCandlish, Kamal Ndousse, Oliver Rausch, Nicholas Schiefer, Da~Yan, Miranda Zhang, and Ethan Perez.
\newblock Towards understanding sycophancy in language models.
\newblock In \emph{International Conference on Learning Representations}, 2024.

\bibitem[Shiiku et~al.(2026)Shiiku, Marjieh, Griffiths, Anglada-Tort, and Jacoby]{shiiku2026hybrid}
Shota Shiiku, Raja Marjieh, Thomas~L. Griffiths, Manuel Anglada-Tort, and Nori Jacoby.
\newblock Hybrid human--{AI} societies balance creativity and diversity.
\newblock PsyArXiv preprint, 2026.
\newblock URL \url{https://osf.io/preprints/psyarxiv/wzh2x_v1}.

\bibitem[Soffer et~al.(2026)Soffer, Shwartz-Ziv, and Shani]{soffer2026trust}
Liron Soffer, Ravid Shwartz-Ziv, and Chen Shani.
\newblock {LLMs} trust their own: Identity-dependent conformity in multi-agent systems.
\newblock arXiv:2609.33495, 2026.
\newblock URL \url{https://arxiv.org/abs/2609.33495}.

\bibitem[Song et~al.(2026)Song, Pala, Zhou, Jin, Zadeh, Li, Herremans, and Poria]{song2026rapport}
Maojia Song, Tej~Deep Pala, Ruiwen Zhou, Weisheng Jin, Amir Zadeh, Chuan Li, Dorien Herremans, and Soujanya Poria.
\newblock Measuring and mitigating rapport bias of large language models under multi-agent social interactions.
\newblock In \emph{International Conference on Learning Representations}, volume 2026, pages 7066--7099, 2026.

\bibitem[{Team Suno}(2025)]{suno}
{Team Suno}.
\newblock Introducing v4.5.
\newblock \url{https://suno.com/blog/introducing-v4-5}, 2025.

\bibitem[Van~de Rijt(2022)]{vanderijt2019self}
Arnout Van~de Rijt.
\newblock Self-correcting dynamics in social influence processes.
\newblock In \emph{Handbook of Sociological Science}, pages 446--473. Edward Elgar Publishing, 2022.

\bibitem[van~de Rijt et~al.(2014)van~de Rijt, Kang, Restivo, and Patil]{vanderijt2014field}
Arnout van~de Rijt, Soong~Moon Kang, Michael Restivo, and Akshay Patil.
\newblock Field experiments of success-breeds-success dynamics.
\newblock \emph{Proceedings of the National Academy of Sciences}, 111\penalty0 (19):\penalty0 6934--6939, 2014.
\newblock \doi{10.1073/pnas.1316836111}.

\bibitem[Weck et~al.(2024)Weck, Manco, Benetos, Quinton, Fazekas, and Bogdanov]{weck2024muchomusic}
Benno Weck, Ilaria Manco, Emmanouil Benetos, Elio Quinton, György Fazekas, and Dmitry Bogdanov.
\newblock Muchomusic: Evaluating music understanding in multimodal audio-language models.
\newblock In \emph{Proceedings of the 25th International Society for Music Information Retrieval Conference (ISMIR)}, 2024.

\bibitem[Weng et~al.(2025)Weng, Chen, and Wang]{weng2025conformity}
Zhiyuan Weng, Guikun Chen, and Wenguan Wang.
\newblock Do as we do, not as you think: {The} conformity of large language models.
\newblock In \emph{International Conference on Learning Representations}, 2025.

\bibitem[Wu et~al.(2025)Wu, Zhao, Chen, Zhou, Lian, and Xie]{wu2025exploring}
Weidong Wu, Qinlin Zhao, Hao Chen, Lexin Zhou, Defu Lian, and Hong Xie.
\newblock Exploring the choice behavior of large language models.
\newblock In Wanxiang Che, Joyce Nabende, Ekaterina Shutova, and Mohammad~Taher Pilehvar, editors, \emph{Findings of the {A}ssociation for {C}omputational {L}inguistics: {ACL} 2025}, pages 5194--5214, Vienna, Austria, July 2025. Association for Computational Linguistics.
\newblock ISBN 979-8-89176-256-5.
\newblock \doi{10.18653/v1/2025.findings-acl.270}.
\newblock URL \url{https://aclanthology.org/2025.findings-acl.270/}.

\bibitem[Wynn et~al.(2025)Wynn, Satija, and Hadfield]{wynn2025talk}
Andrea Wynn, Harsh Satija, and Gillian Hadfield.
\newblock Talk isn't always cheap: Understanding failure modes in multi-agent debate.
\newblock arXiv:2509.05396, 2025.
\newblock URL \url{https://arxiv.org/abs/2509.05396}.

\bibitem[Ye et~al.(2025)Ye, Wang, Huang, Chen, Zhang, Moniz, Gao, Geyer, Huang, Chen, et~al.]{ye2025justice}
Jiayi Ye, Yanbo Wang, Yue Huang, Dongping Chen, Qihui Zhang, Nuno Moniz, Tian Gao, Werner Geyer, Chao Huang, Pin-Yu Chen, et~al.
\newblock Justice or prejudice? quantifying biases in llm-as-a-judge.
\newblock In \emph{International Conference on Learning Representations}, volume 2025, pages 102351--102390, 2025.

\bibitem[Zhang et~al.(2024)Zhang, Xu, Zhang, Liu, Hooi, and Deng]{zhang2024exploring}
Jintian Zhang, Xin Xu, Ningyu Zhang, Ruibo Liu, Bryan Hooi, and Shumin Deng.
\newblock Exploring collaboration mechanisms for {LLM} agents: A social psychology view.
\newblock In Lun-Wei Ku, Andre Martins, and Vivek Srikumar, editors, \emph{Proceedings of the 62nd Annual Meeting of the {A}ssociation for {C}omputational {L}inguistics (Volume 1: Long Papers)}, pages 14544--14607, Bangkok, Thailand, August 2024. Association for Computational Linguistics.
\newblock \doi{10.18653/v1/2024.acl-long.782}.
\newblock URL \url{https://aclanthology.org/2024.acl-long.782/}.

\bibitem[Zheng et~al.(2024)Zheng, Zhou, Meng, Zhou, and Huang]{zheng2024robust}
Chujie Zheng, Hao Zhou, Fandong Meng, Jie Zhou, and Minlie Huang.
\newblock Large language models are not robust multiple choice selectors.
\newblock In \emph{The Twelfth International Conference on Learning Representations}, 2024.
\newblock URL \url{https://openreview.net/forum?id=shr9PXz7T0}.

\bibitem[Zheng et~al.(2023)Zheng, Chiang, Sheng, Zhuang, Wu, Zhuang, Lin, Li, Li, Xing, Zhang, Gonzalez, and Stoica]{zheng2023judging}
Lianmin Zheng, Wei-Lin Chiang, Ying Sheng, Siyuan Zhuang, Zhanghao Wu, Yonghao Zhuang, Zi~Lin, Zhuohan Li, Dacheng Li, Eric~P. Xing, Hao Zhang, Joseph~E. Gonzalez, and Ion Stoica.
\newblock Judging {LLM}-as-a-judge with {MT-Bench} and {Chatbot Arena}.
\newblock In \emph{Advances in Neural Information Processing Systems}, volume~36, pages 46595--46623, 2023.

\bibitem[Zhong et~al.(2025)Zhong, Fang, Shi, Huang, Zheng, Du, Chen, and Wang]{zhong2025spiral}
Mingze Zhong, Meng Fang, Zijing Shi, Yuxuan Huang, Shunfeng Zheng, Yali Du, Ling Chen, and Jun Wang.
\newblock Spiral of silence in large language model agents.
\newblock In Christos Christodoulopoulos, Tanmoy Chakraborty, Carolyn Rose, and Violet Peng, editors, \emph{Findings of the {A}ssociation for {C}omputational {L}inguistics: {EMNLP} 2025}, pages 23238--23253, Suzhou, China, November 2025. Association for Computational Linguistics.
\newblock ISBN 979-8-89176-335-7.
\newblock \doi{10.18653/v1/2025.findings-emnlp.1262}.
\newblock URL \url{https://aclanthology.org/2025.findings-emnlp.1262/}.

\bibitem[Zhu et~al.(2025)Zhu, Zhang, Stafford, Collier, and Vlachos]{zhu2025conformity}
Xiaochen Zhu, Caiqi Zhang, Tom Stafford, Nigel Collier, and Andreas Vlachos.
\newblock Conformity in large language models.
\newblock In \emph{Proceedings of the 63rd Annual Meeting of the Association for Computational Linguistics (Volume 1: Long Papers)}, pages 3854--3872, 2025.

\end{thebibliography}
\bibliographystyle{plainnat}

%%%%%%%%%%%%%%%% START OF SUPPLEMENT %%%%%%%%%%%%%%%

\clearpage
\appendix

% Figures, tables, equations and pages in the supplement are numbered S1, S2 etc.
\renewcommand{\thefigure}{S\arabic{figure}}
\renewcommand{\thetable}{S\arabic{table}}
\renewcommand{\theequation}{S\arabic{equation}}
\renewcommand{\thepage}{S\arabic{page}}
\setcounter{figure}{0}
\setcounter{table}{0}
\setcounter{equation}{0}
\setcounter{page}{1}

%%%%%%%%%%%%%%%% MATERIALS AND METHODS %%%%%%%%%%%%%%%

\section{Related Work}\label{app:related}
 
\subsection{Conformity and sycophancy in individual LLMs}
 
A large literature documents how individual LLMs shift their responses toward
opinions presented in their input. Studies of sycophancy found that agreement
with users increased with model size \citep{perez2023discovering}, appeared
across free-form tasks and was favored by human preference judgments
\citep{sharma2024towards}, and led models to abandon correct answers after
user rebuttals \citep{fanous2025syceval}. A second line of work adapted the
Asch paradigm by displaying peer answers before the model responds. Weng et al.\
introduced five interaction protocols, including multi-round histories of
correct guidance followed by incorrect guidance; all eleven models exhibited
conformity, while larger models were more independent
\citep{weng2025conformity}. Zhu et al.\ found that conformity increased with
the number and confident tone of unanimous confederates and was predicted by
the model's uncertainty \citep{zhu2025conformity}. Other studies linked
conformity to group size and task difficulty \citep{bellina2026conformity},
peers' group identity \citep{soffer2026trust}, prior agreement with peers
\citep{song2026rapport}, and relative confidence between the model and its
peers \citep{cho2025herd}. Models were easier to mislead than to correct:
wrong peer agreement increased harmful revision more than correct agreement
increased beneficial revision \citep{qu2026easier}. Across 23 models,
mitigations generally traded resistance to incorrect peers against receptivity
to correct ones; reasoning improved both on derivable MMLU tasks, but the
trade-off persisted on GPQA and SimpleQA \citep{hussain2026conformity}.
Popularity cues also influenced LLM judges \citep{ye2025justice} and product
recommendations \citep{filandrianos2025bias}.
 
These studies primarily examine how individual models respond to controlled
social information, rather than how successive decisions update a cumulative
popularity signal. Our study asks how individual susceptibility develops
into population-level cascades when recommendations generate the signal seen
by subsequent agents. The outcomes are divergence from independent judgment,
unpredictability across replicate worlds, and self-correction, rather than
whether a model adopts a supplied answer.
 
\subsection{Collective dynamics in interacting LLM populations}
 
When agents respond to one another, social influence becomes endogenous.
Multi-agent debate improved factuality \citep{du2024improving}, but also
produced conformity \citep{zhang2024exploring,baltaji2024conformity},
deference to larger or more capable groups \citep{choi2025empirical}, and
accuracy losses when agents abandoned correct answers \citep{wynn2025talk}.
In a seven-agent misinformation task, denser networks accelerated convergence
but raised the risk of confident incorrect consensus \citep{han2026conformity}.
Larger populations developed shared conventions, collective biases, and
minority-driven tipping \citep{flintashery2025emergent}, reached consensus
or fragmented depending on prompted bias \citep{chuang2024simulating}, and
became trapped in states opposed to their intrinsic preferences
\citep{demarzo2026conformity}. Forum simulations found greater conformity
in smaller models and greater resistance in reasoning-optimized models
\citep{lin2025towards}. Observational analyses of Chirper.ai, a platform
populated by LLM agents, documented homophily and social influence among
approximately 32,000 agents \citep{hashemi2026empirical,hashemi2025collective}.
 
Sequential decisions and accumulating social signals provide closer
comparisons. Jain and Krishnamurthy combined LLM outputs with Bayesian
decision mechanisms to study social learning and information cascades
\citep{jain2025interacting}. Zhong et al.\ had 100 agents rate a film in
sequence while observing earlier agents' mean rating, producing anchoring
and, with personas, majority dominance \citep{zhong2025spiral}. In a
simulated microblogging platform, responses to endogenous popularity counts
varied with information load and normative context \citep{peng2026engagement}.
Wu et al.\ studied choices on a question-answering platform with independent
conditions, accumulating engagement counts, and experimentally seeded
popularity, finding reinforcement of existing preferences and shifts toward
socially favored alternatives \citep{wu2025exploring}.
 
Our study extends this work by connecting individual susceptibility
to the distribution, unpredictability, and self-correction of collective
recommendations in one controlled design. Agents act once in sequence and
choose among fixed alternatives whose intrinsic appeal is measured
independently. We combine this design with a fixed selection budget, separate
selection and endorsement stages, and a manipulation of whether popularity
also determines list position. Separating selection from endorsement makes
attention a bottleneck: an item must first be selected before it can receive
a recommendation. Popularity can therefore reinforce both exposure and
evaluation. Comparing 13 models tests whether greater individual capability
is associated with collective robustness. Replicated worlds reveal whether
early choices redirect collective outcomes, while temporal comparisons test
whether accumulating recommendations restore alignment with independent appeal.
A human benchmark places the magnitude of these effects in context.
 
\subsection{Hybrid human--AI societies and feedback loops}
 
Human--AI interaction can amplify initially small judgment biases
\citep{glickman2025human}. In story-revision networks, AI-only populations
lost diversity whereas hybrid populations better balanced creativity and
diversity \citep{shiiku2026hybrid}; opinion-diffusion experiments reported
preliminary evidence of lower final polarization in hybrid networks
\citep{gaubert2026experimental}. Recommender-system research likewise shows
that feedback can homogenize choices and amplify popularity bias
\citep{fleder2009blockbuster,chaney2018algorithmic,mansoury2020feedback}.
Our experiment isolates a loop operating through cumulative recommendations,
without changing model weights or the alternatives available. It tests how
the population's own choices reshape subsequent decisions within homogeneous
LLM populations; extending this design to mixed human--AI populations would
test whether population composition changes cascade strength and correction.
 
\section{Materials and Methods}\label{app:methods}

The multiple-worlds design, population sizes, and influence conditions are
described in the main text. The following sections provide additional design
details, implementation procedures, and statistical methods.

\subsection{Song stimuli}

Lyrics ranged from 131 to 172 words, with a mean length of 153.8 words.
Table~\ref{tab:song-lengths} lists all 48 titles and their lyric lengths.

\begin{table}[H]
\centering
\caption{\textbf{Song stimuli and lyric length.} Word counts were computed
using whitespace-separated tokens, including section markers.}
\label{tab:song-lengths}
\begin{tabular}{lr@{\hspace{1.5cm}}lr}
\toprule
Title & Words & Title & Words \\
\midrule
Map Beneath the Stars & 172 & Briefly Burning & 156 \\
Voice Like Silk & 158 & Electric Soulmates & 155 \\
Neon Garden & 137 & Concrete Undertow & 145 \\
Uncaught Thoughts & 165 & Plastic Moons & 150 \\
Where the Day Gives Up & 158 & Clouds in My Tea & 137 \\
Drift Until It Ends & 144 & Stealing Stars & 162 \\
Broken Clock & 172 & Fractured Beams & 150 \\
Pixelated Dreams & 159 & Helium Strings & 158 \\
Edge of the Unknown & 161 & Parking Lot Marquee & 140 \\
Running Uphill at Midnight & 162 & Someone Else's Name & 162 \\
Paint the Silence & 170 & Steering Clear & 152 \\
Golden Streams & 154 & The Sun Borrowed My Smile & 149 \\
Before the Day & 161 & On the Fire Escape & 142 \\
Digital Driftwood & 153 & Friend to the Lost & 167 \\
Velvet Web & 168 & Sparks in the Power Lines & 141 \\
Through the Cracks & 133 & Falling Satellites & 142 \\
Compass Ink & 160 & Colors I Could Be & 138 \\
The Alley Understands & 164 & Soaked and Torn & 142 \\
Voltage Reverie & 148 & Not the Sailor & 154 \\
Midnight Conversations & 172 & Where Whispers Go & 162 \\
Lazy Eye on the Tide & 156 & Moonlight Laundry & 131 \\
Matches on Water & 145 & Brick and Mortar Heart & 171 \\
Storm-Folded Letters & 142 & Blue-Light Crowd & 167 \\
Asphalt Symphony & 162 & Cactus in the Rain & 135 \\
\bottomrule
\end{tabular}
\end{table}

\subsection{Persona construction and agent heterogeneity}

Persona conditioning introduced structured variation in the inputs guiding
agents' preferences. With a fixed model, prompt, and stimuli, stochastic
decoding can produce different responses across repeated calls, but these
responses remain draws from the same underlying conditional distribution.
Assigning distinct demographic profiles and musical preferences varied the
conditioning information itself, allowing us to study social feedback in a
population with heterogeneous preference prompts.

Persona conditioning is widely used to construct heterogeneous LLM populations
by conditioning model responses on individual characteristics
\citep{argyle2023out, aher2023using, ashokkumar2026large}. Prior work has also
found that persona prompting increased the diversity of model responses in a
systematic and repeatable manner \citep{frohling-etal-2025-personas}. Each
persona was assigned a primary and secondary preferred music genre drawn from
the following: Classical, Jazz, Blues, Rock, Pop, Folk, Country,
Gospel/Religious, Hip-Hop/Rap, R\&B/Soul, Electronic/Dance, Reggae, Latin,
Metal, and World/Global. The persona also included age, gender, and country of
residence, sampled to approximate the demographic composition of participants
in the original Music Lab experiment \citep{salganik2006experimental}. However,
we do not claim that individual agents reproduced individual human listeners.
Persona-conditioned synthetic samples can still underrepresent human variation
\citep{santurkar2023whose, bisbee2024synthetic}. The same persona-generating
procedure was used in the independent and influence conditions to avoid
systematic differences in population construction alongside exposure to
recommendation counts.

\subsection{Prompts}

\begin{figure}[H]
    \centering
    \includegraphics[width=\textwidth]{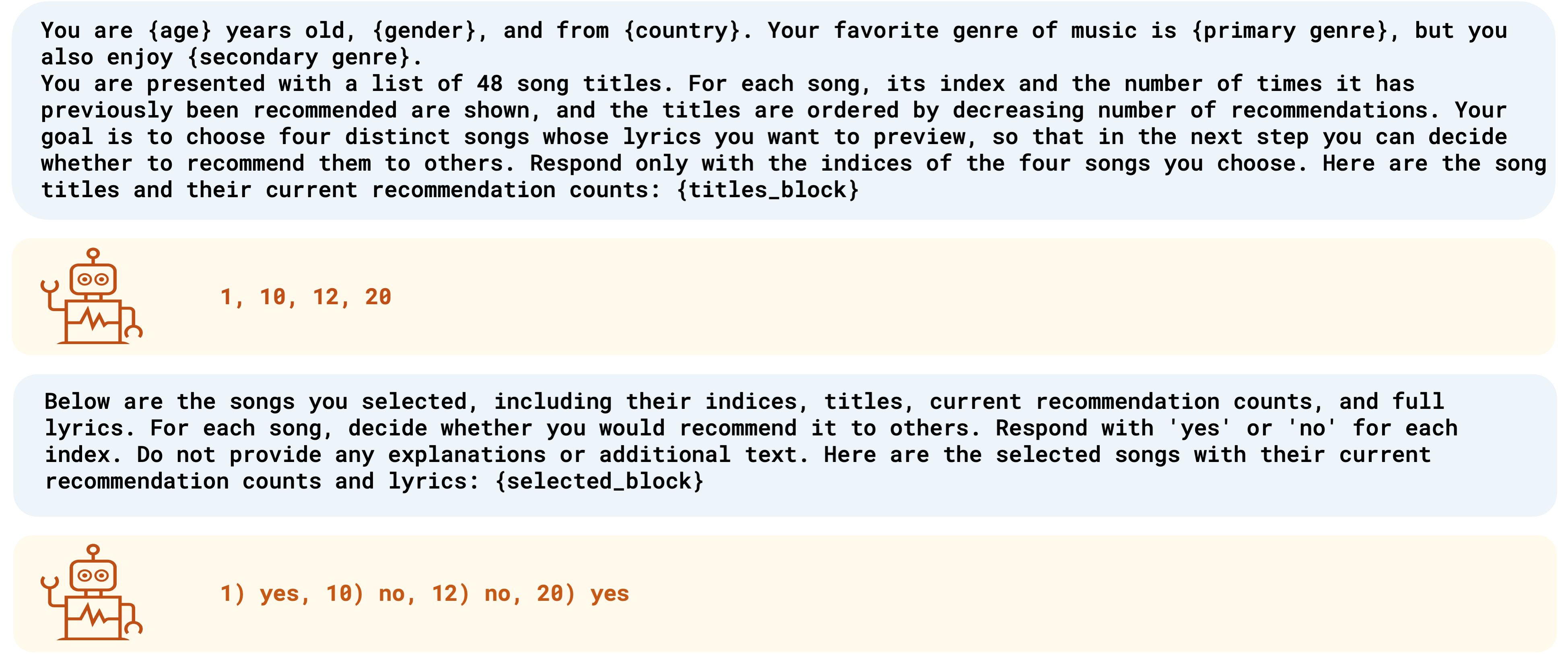}
    \caption{\textbf{Prompts used in the strong-influence condition.} The upper
    blue box contains the prompt for the selection stage and the lower blue box
    contains the prompt for choosing songs to recommend. The yellow boxes show
    illustrative responses. Braced fields identify persona, song, and
    recommendation-count inputs. Counts reflect the cumulative number of
    recommendations before the agent's decisions and were displayed in both
    stages. The weak-influence condition used the same templates except that
    the clause describing popularity-based ordering was omitted and title
    order was randomized independently for each agent.}
    \label{fig:prompt_influence}
\end{figure}

\begin{figure}[H]
    \centering
    \includegraphics[width=\textwidth]{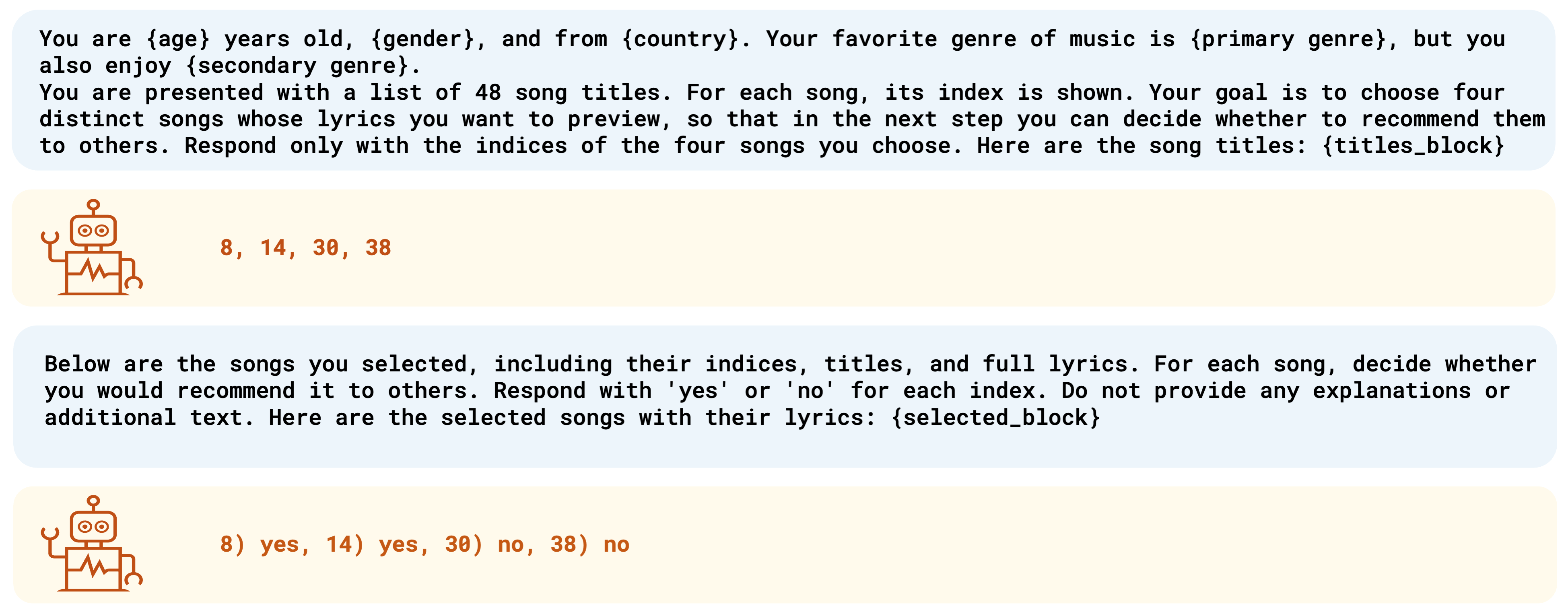}
    \caption{\textbf{Prompts used in the independent condition.} The upper and
    lower blue boxes contain selection- and evaluation-stage prompt templates,
    with illustrative responses in the yellow boxes. Persona framing, task
    structure, and response format matched the influence conditions, but
    recommendation counts and references to earlier choices were omitted. Title
    order was randomized independently for each agent.}
    \label{fig:prompt_independent}
\end{figure}

Figs.~\ref{fig:prompt_influence} and \ref{fig:prompt_independent} provide
the complete prompt templates. All prompts began with the agent's
age, gender, country of residence, and primary and secondary preferred music
genres. No conversational history was shared across agents. Apart from the count and ordering manipulations described in the
main text, prompt structure was held constant across conditions. The clause
describing popularity-based ordering was omitted from the weak-influence
selection prompt. Agents returned four song indices in the selection task and
a yes or no decision for each selected song in the recommendation task,
without additional comments or explanation. These constrained outputs
standardized parsing across models and conditions; for reasoning models, the
returned answers were constrained but not the reasoning process preceding them.

\subsection{Models}

Non-reasoning models used a sampling temperature of 1.0. Reasoning models used
medium reasoning effort where an effort level was supported. Within each
model and experimental condition, all agents used identical inference settings,
and other generation parameters were retained at their provider defaults unless
otherwise specified.

\subsection{Human benchmark and design differences}\label{app:human}

Human benchmark values were computed from the original Music Lab data
\citep{salganik2006experimental} using the same outcome definitions as for the
LLM populations, with downloads in place of recommendations. Weak influence
corresponded to experiment 1 and strong influence to experiment 2. Each human
experiment included one independent world and eight influence worlds, with
its own independent world serving as the baseline. The self-correction
diagnostic was applied to the same data following van de Rijt
\citep{vanderijt2019self}.

Our design preserved the causal structure of Music Lab: sequential choices,
parallel worlds that began from zero and evolved separately, an independent
world without social information, and two influence conditions that differed
in whether popularity also determined display position. The implementation was
adapted to LLM agents in six ways, each discussed below.

\paragraph{Newly generated songs.}
Music Lab used songs by unknown bands so that participants would not know the
songs or their popularity. For LLMs, reusing the original songs risked prior
exposure during model training, which could encode not only familiarity with a
song but also information about its popularity and reception. Newly generated
songs reduced the risk that such an external popularity signal would
contaminate the endogenous recommendation counts as the source of influence.

\paragraph{Lyrics rather than audio.}
We considered presenting audio to multimodal models but selected a lyrics-only
design for three reasons. First, audio would have required audio-capable
models, a relatively small class less likely to be used in multi-agent
systems, and the computational cost of audio evaluation for over 160,000
agents was a practical constraint. Text allowed identical materials to be
presented to all 13 models and aligned the experiment with language-based
judgments in recommender systems, automated evaluation, and multi-agent
systems. Second, audio input does not ensure that judgments depend on musical
content: in a previous study, three of five audio-language models showed no
significant performance decline when the original audio was replaced with
noise or an unrelated track, consistent with substantial reliance on textual
cues \citep{weck2024muchomusic}. Third, even with newly generated audio, the
stimuli would still differ from the original songs. We therefore leave cascade
dynamics in multimodal populations to future research.

\paragraph{Recommendation rather than download.}
Music Lab participants rated each song while listening and then decided
whether to download it. Downloading has no direct behavioral counterpart for an
agent, whereas recommendation is a widely used LLM task. The substitution
preserved the feedback mechanism: each affirmative choice increased an
observable count of prior choices and thereby changed the information
encountered by subsequent agents.

\paragraph{Fixed selection budget.}
Music Lab participants chose how many songs to sample. We fixed the budget at
four, approximating the average number sampled by participants (3.8 and 3.6 in
experiments 1 and 2, respectively) \citep{salganik2006experimental}, to
standardize evaluation opportunities across agents and models; the main text
describes how this budget reflects candidate-selection bottlenecks in LLM
systems. Increasing the budget to 12 reduced divergence and unpredictability
but preserved the main patterns (figs.~\ref{fig:selection_kl_robustness} and
\ref{fig:selection_u_robustness}).

\paragraph{List rather than grid.}
Music Lab displayed songs in a randomized grid in experiment 1 and in a
single-column list in experiment 2, ordered by downloads in the influence
worlds and randomly in the independent world \citep{salganik2006experimental}.
A list is the natural text representation for LLM input, so we used a
randomized list in the independent and weak-influence conditions and a
popularity-ordered list under strong influence. The weak-influence condition
therefore preserved the defining feature of the human weak-influence
condition, popularity information without popularity-based ordering, but not
its interface. For the same reason, human position bias under weak influence
is not directly comparable with that of LLM agents and is not reported
(fig.~\ref{fig:position_bias}).

\paragraph{Shared independent baseline.}
Each Music Lab experiment had its own independent world. Because both LLM
influence conditions used the same model, stimuli, persona-generating
procedure, and inference settings, we compared all 16 influence worlds with a
single independent world per model, so the weak- and strong-influence
conditions shared a common baseline.

These differences limit attribution of human--LLM contrasts to population
type, so we treat the human results as a qualitative benchmark. Comparisons
among LLMs used the same stimuli, task, and condition definitions.

\subsection{Outcome measures}\label{app:measures}

We measured market share using the distribution of song recommendation counts
in each world. Let $K=48$ denote the number of songs, $r_i^{(w)}$ the number of
recommendations received by song $i$ in world $w$, and
$R^{(w)}=\sum_{i=1}^{K}r_i^{(w)}$ the total number of recommendations in that
world. Following the original Music Lab experiment
\citep{salganik2006experimental}, the market share of song $i$ was
\begin{equation}
p_{w}(i) = \frac{r_{i}^{(w)}}{R^{(w)}},
\qquad
\sum_{i=1}^{K}p_{w}(i) = 1.
\end{equation}
Market shares accounted for differences in total recommendation counts because
agents could recommend between zero and four songs. Independent-condition
unpredictability and inequality were estimated through resampling as described
below.

\paragraph{KL divergence.}

Kullback--Leibler (KL) divergence compares each song's market share under
influence with its independent share, measuring displacement from the
population-specific baseline. For influence world $w$, we calculated
\begin{equation}
D_{\mathrm{KL}}^{(w)} = \sum_{i=1}^{K}\widetilde{p}_{w}(i)
\log\frac{\widetilde{p}_{w}(i)}{\widetilde{p}_{\mathrm{ind}}(i)},
\end{equation}
where $\widetilde{p}$ denotes smoothed market shares. Divergence equals zero
when the two smoothed distributions coincide, and larger values indicate
greater departure from the independent baseline. We reported the arithmetic mean
over the eight worlds in each influence condition.

To avoid infinite divergence when a song had positive influence-world share but
zero independent-world share, we applied Jeffreys smoothing to both
distributions by adding a pseudocount of $\alpha=0.5$ to every song before
normalization:
\begin{equation}
\widetilde{p}_{w}(i) = \frac{r_{i}^{(w)} + 0.5}{R^{(w)} + 0.5K}.
\end{equation}
The same procedure was applied to the independent distribution. Jeffreys
smoothing was used only for KL divergence; all remaining measures used
unsmoothed market shares or recommendation counts. Unlike Gini inequality,
which measures concentration within a world, KL divergence also distinguishes
which songs received larger shares relative to the independent baseline.

\paragraph{Unpredictability.}

Unpredictability measured the extent to which parallel worlds containing the
same songs and LLM model nevertheless evolved toward different collective
outcomes. Following Salganik, Dodds, and Watts
\citep{salganik2006experimental}, the unpredictability of song $i$ was the
average absolute difference in its market share across all pairs of influence
worlds:
\begin{equation}
u_{i} = \binom{W}{2}^{-1}\sum_{j=1}^{W-1}\sum_{k=j+1}^{W}
\left| p_{j}(i) - p_{k}(i) \right|,
\end{equation}
where $W=8$ is the number of worlds in each influence condition. Overall
unpredictability was the average across all $K=48$ songs:
\begin{equation}
U = \frac{1}{K}\sum_{i=1}^{K}u_{i}.
\end{equation}
Larger values indicate greater variation in collective outcomes across
worlds. Increased unpredictability under influence relative to independent
evaluation indicates greater dependence on the history through which
recommendations accumulated.

With eight influence worlds, each song's unpredictability was based on
$\binom{8}{2}=28$ world pairs. %These pairs shared worlds and were not independent experimental replicates.
For the independent condition, we followed the original Music Lab procedure: we
randomly split the independent population into two equal subpopulations of 720
agents, matching the size of an influence world, calculated the mean absolute
difference in market shares between them, and averaged the results over 1,000
random splits \citep{salganik2006experimental}.

\paragraph{Relative unpredictability.}

The capability analysis divided each model's influence-condition
unpredictability by its own independent baseline:
\begin{equation}
U_{\mathrm{rel}} = \frac{U_{\mathrm{inf}}}{U_{\mathrm{ind}}},
\end{equation}
where $U_{\mathrm{inf}}$ is unpredictability in the specified influence
condition and $U_{\mathrm{ind}}$ is the independent estimate obtained by
repeated splitting for the same model. A ratio of one indicates
equal variability, while a ratio above one indicates proportional
amplification under influence. 

\paragraph{Pearson correlation between worlds.}

We measured the extent to which popularity reflected intrinsic appeal as the
Pearson correlation between market shares in the independent and influence
worlds. For each population and influence condition, we pooled the 48 songs
across the eight influence worlds, yielding $8\times48=384$ paired
observations. Each influence-world market share was paired with the
corresponding song's market share in the independent world. Let $x_{\ell}$
denote the independent-world market share and $y_{\ell}$ the corresponding
influence-world market share for pooled observation $\ell$. Pearson's
correlation was
\begin{equation}
r = \frac{\sum_{\ell=1}^{KW}\left( x_{\ell} - \overline{x} \right)\left( y_{\ell} - \overline{y} \right)}
{\sqrt{\sum_{\ell=1}^{KW}\left( x_{\ell} - \overline{x} \right)^{2}}
\sqrt{\sum_{\ell=1}^{KW}\left( y_{\ell} - \overline{y} \right)^{2}}}.
\end{equation}
Higher positive correlations indicate that songs with greater intrinsic appeal
acquired larger market shares under influence. The 384 pairs were not 384
independent replicates: each song's independent share was reused in eight
pairs, songs within a world shared the same feedback process, and their shares
summed to one. A conventional Pearson test treating these pairs as independent
would therefore give misleading significance levels. Accordingly,
table~\ref{tab:pearson_results} reports descriptive coefficients without
significance tests.

\paragraph{Correlation between early movers and later agents.}
 
Adapting the analysis of early movers in a multiple-worlds study of partisan
opinion \citep{macy2019opinion}, we also measured how well the choices of the
first agents in each influence world predicted the choices of the agents who
followed them. Within each influence world, we calculated each song's market
share among the recommendations made by the first 30 agents and its market
share among the recommendations made by all later agents, excluding the first
30 agents' recommendations from the latter. As for the correlation between
worlds, we pooled the 48 songs across the eight influence worlds and calculated
the Pearson correlation between the two sets of shares. A high correlation
indicates that later agents largely reproduced the ranking established by the
first agents. Comparing this correlation with the correlation between influence
and independent worlds indicates whether later popularity was better predicted
by early movers or by intrinsic appeal. For the human benchmark, we used the
first 30 participants in each Music Lab influence world. 
 
\begin{table}[H]
\centering
\caption{\textbf{Correspondence of popularity with intrinsic appeal and with early movers in the
social influence conditions.}  The first two columns report Pearson correlations between
song market shares under social influence and in the independent condition,
pooling the 48 songs across the eight influence worlds (384 paired song--world
observations per population and condition). Higher values indicate greater
alignment with intrinsic appeal. The last two columns report Pearson
correlations between market shares among the first 30 agents (participants for
humans) in each influence world and market shares among all later agents in
the same world, pooled in the same way; higher values indicate that later
agents reproduced the ranking established by early movers. Class
means are arithmetic averages of model-level coefficients.}
\label{tab:pearson_results}
\begin{tabular}{lcccc}
\toprule
 & \multicolumn{2}{c}{Influence vs.\ independent} & \multicolumn{2}{c}{First 30 vs.\ later agents} \\
\cmidrule(lr){2-3}\cmidrule(lr){4-5}
Population & Weak & Strong & Weak & Strong \\
\midrule
Human data          & 0.740 & 0.557 & 0.316 & 0.723 \\
\midrule
GPT-5 nano          & 0.533 & 0.389 & 0.844 & 0.963 \\
GPT-5               & 0.885 & 0.815 & 0.823 & 0.889 \\
Claude Sonnet 4.6   & 0.741 & 0.779 & 0.947 & 0.858 \\
Gemini 3 Flash      & 0.860 & 0.831 & 0.889 & 0.857 \\
DeepSeek-R1         & 0.672 & 0.633 & 0.888 & 0.844 \\
GPT-4o mini         & 0.349 & 0.149 & 0.925 & 0.986 \\
GPT-4o              & 0.666 & 0.673 & 0.765 & 0.792 \\
Claude Haiku 4.5    & 0.812 & 0.674 & 0.870 & 0.872 \\
Gemma 3 4B          & 0.305 & 0.118 & 0.792 & 0.998 \\
Gemma 3 12B         & 0.684 & 0.425 & 0.866 & 0.976 \\
Gemma 3 27B         & 0.341 & 0.177 & 0.979 & 0.982 \\
DeepSeek-V3.1       & 0.375 & 0.317 & 0.758 & 0.876 \\
Qwen3-Next 80B      & 0.365 & 0.151 & 0.972 & 0.970 \\
\midrule
Mean reasoning     & 0.738 & 0.689 & 0.878 & 0.882 \\
Mean non-reasoning & 0.487 & 0.336 & 0.866 & 0.932 \\
\bottomrule
\end{tabular}
\end{table}

\paragraph{Self-correction toward intrinsic appeal.}

Cascades can self-correct if intrinsic appeal is sufficiently strong relative
to social influence \citep{vanderijt2019self}. 
Following van de Rijt \citep{vanderijt2019self}, we divided each influence
world's chronological sequence of recommendation events into two approximately
equal halves. Let $q_i$ denote the total recommendations received by song $i$
in the corresponding independent world, and $r_{i,1}^{(w)}$ and
$r_{i,2}^{(w)}$ its counts within the first and second halves of influence
world $w$. Second-half counts excluded first-half recommendations. We
calculated the Spearman correlation across the 48 songs:
\begin{equation}
\rho_{1}^{(w)} = \rho_{S}\left( \left( r_{i,1}^{(w)} \right)_{i=1}^{K},
\left( q_{i} \right)_{i=1}^{K} \right),
\qquad
\rho_{2}^{(w)} = \rho_{S}\left( \left( r_{i,2}^{(w)} \right)_{i=1}^{K},
\left( q_{i} \right)_{i=1}^{K} \right),
\end{equation}
where 1 and 2 index the first and second halves of the sequence of submitted
recommendations. Self-correction was indicated by the change in correlation
between halves:
\begin{equation}
\Delta\rho^{(w)} = \rho_{2}^{(w)} - \rho_{1}^{(w)}.
\end{equation}
Positive values indicate that popularity became more closely aligned with
intrinsic appeal as recommendations accumulated; values near zero indicate
little net change, and negative values indicate movement away from the
independent ranking. For each population, we report the mean change across the
eight influence worlds and the number of worlds in which $\Delta\rho^{(w)}>0$.
We tested the directional hypothesis of positive change using one-sided sign
and Wilcoxon signed-rank tests across the eight worlds.

\paragraph{Market-share inequality.}

Following the original Music Lab experiment \citep{salganik2006experimental},
we used the Gini coefficient to measure inequality of market shares in each
world:
\begin{equation}
G_{w} = \frac{1}{2K}\sum_{i=1}^{K}\sum_{j=1}^{K}\left| p_{w}(i) - p_{w}(j) \right|.
\end{equation}
The coefficient equals zero for equal shares and increases with concentration,
with a maximum of $(K-1)/K$. For the LLM independent condition, we drew 1,000
bootstrap samples of 720 agents, calculated the Gini coefficient for each
sample, and reported their arithmetic mean. We compared mean inequality across
the eight influence worlds with this independent baseline. (See Supplementary
results below.)

\subsection{Robustness checks}

We repeated the strong-influence experiment with temperature reduced from
1.0 to 0.7 for the eight non-reasoning models, and separately with the
selection budget increased from four to 12 songs for all 13 models. Each
alternative setting included a matched independent world, so comparisons used
baselines generated under the same temperature or selection budget. All other
procedures were unchanged. The results are reported in
figs.~\ref{fig:temperature_robustness} to \ref{fig:selection_u_robustness}.

%%%%%%%%%%%%%%%% SUPPLEMENTARY RESULTS %%%%%%%%%%%%%%%

\FloatBarrier
\section{Supplementary Results}

\subsection{Model intelligence and susceptibility to social influence}

Fig.~\ref{fig:capability_u} in the main text reports the association between
outcome measures and an established measure of model intelligence -- the
Artificial Analysis Intelligence Index v4.1.1 \citep{artificialanalysis}. This
continuous measure provides a more refined analysis compared to the binary
classification of reasoning and non-reasoning models. These associations are
descriptive, not causal effects of intelligence, and do not control for any of
several factors that affect model performance on the AAII task inventory,
including reasoning class, scale, architecture, training, and post-training.
Reasoning models have a mean AAII score of 32.0, compared with 11.4 among
non-reasoning models (table~\ref{tab:models}). The classes nevertheless
overlapped: GPT-5 nano and DeepSeek-R1 both scored 19, below the non-reasoning
Claude Haiku 4.5 (24) and DeepSeek-V3.1 (21). The continuous index therefore
captured variation beyond the binary classification.

\begin{table}[H]
\centering
\caption{\textbf{AAII scores by reasoning class.} Models were classified by
developer designation and the inference mode used in the experiments.
Artificial Analysis Intelligence Index v. 4.1.1 scores were used in the
analyses \citep{artificialanalysis}. Higher scores indicate greater benchmarked
intelligence.}
\label{tab:models}
\begin{tabular}{lllr}
\toprule
Model & Developer & Class & Intelligence Index \\
\midrule
GPT-5             & OpenAI          & Reasoning     & 35 \\
GPT-5 nano        & OpenAI          & Reasoning     & 19 \\
Gemini 3 Flash    & Google DeepMind & Reasoning     & 39 \\
DeepSeek-R1       & DeepSeek        & Reasoning     & 19 \\
Claude Sonnet 4.6 & Anthropic       & Reasoning     & 48 \\
\midrule
GPT-4o           & OpenAI          & Non-reasoning & 11 \\
GPT-4o mini      & OpenAI          & Non-reasoning & 7 \\
Gemma 3 4B       & Google DeepMind & Non-reasoning & 1 \\
Gemma 3 12B      & Google DeepMind & Non-reasoning & 6 \\
Gemma 3 27B      & Google DeepMind & Non-reasoning & 7 \\
DeepSeek-V3.1    & DeepSeek        & Non-reasoning & 21 \\
Claude Haiku 4.5 & Anthropic       & Non-reasoning & 24 \\
Qwen3-Next 80B   & Alibaba         & Non-reasoning & 14 \\
\bottomrule
\end{tabular}
\end{table}

\subsection{Social influence increased inequality of market shares}

\begin{figure}[H]
\centering
\includegraphics[width=\textwidth]{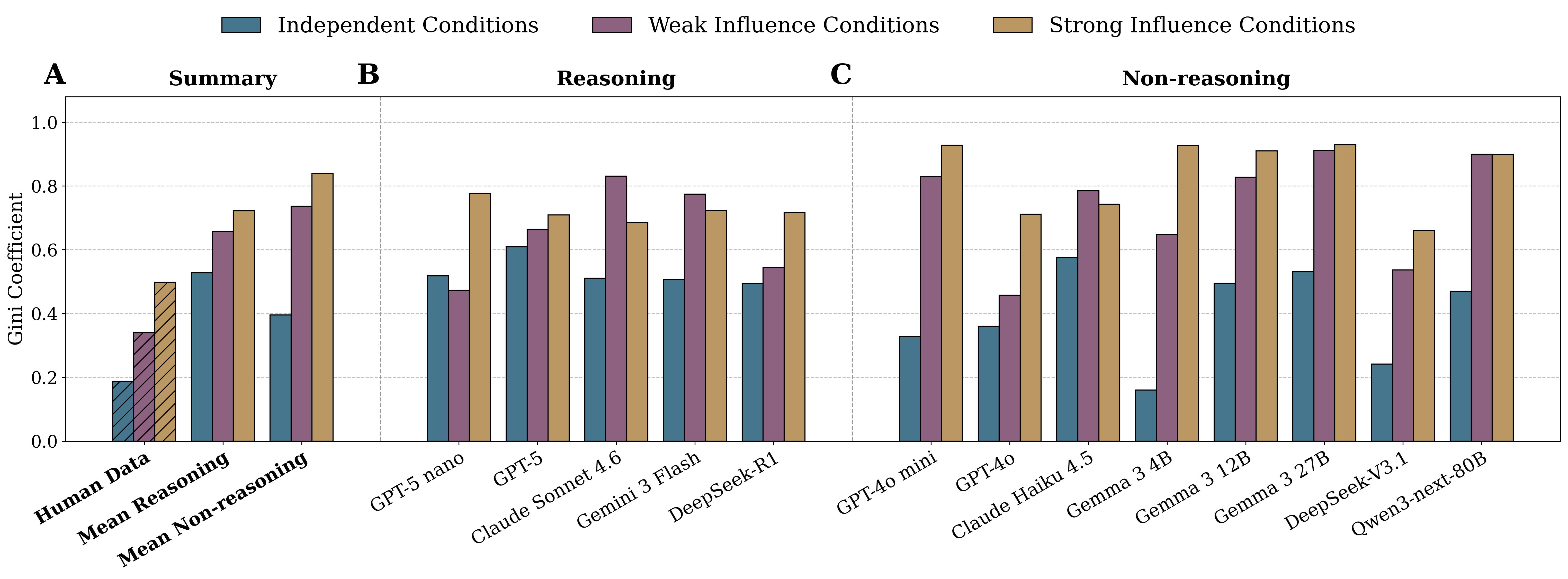}
\caption{\textbf{Social influence generally increased market-share
inequality.} Inequality was measured by the Gini coefficient of market shares
across the 48 songs; larger values indicate greater concentration of popularity
among fewer songs. Gold, purple, and blue denote strong influence, weak
influence, and independent evaluation, respectively. The gold and purple bars
report the Gini coefficient averaged over the eight worlds in each influence
condition. Blue bars show the Gini coefficient averaged over 1,000 bootstrap
samples drawn from the independent world. Each sample was the same size as an
influence world (720 agents for LLMs). Hatched bars show human values from the
original Music Lab data \citep{salganik2006experimental} for comparison with
unweighted class means among the five reasoning and eight non-reasoning models.
\textbf{A.} On average, both model classes were more unequal than humans in the
independent condition. Relative to their independent baselines, reasoning
models showed smaller mean influence-driven increases than humans, whereas
non-reasoning models showed larger increases under both influence conditions.
\textbf{B} and \textbf{C.} Inequality exceeded the independent baseline in all 13 models in the
strong-influence condition and in 12 models under weak influence, with GPT-5
nano the lone exception.}
\label{fig:gini}
\end{figure}

The original Music Lab study used the Gini coefficient to measure inequality in
the distribution of market shares in the independent and influence worlds
\citep{salganik2006experimental}. Inequality captures the concentration
associated with cumulative advantage but not the song-specific displacement of
popularity from intrinsic appeal. We therefore used KL divergence as the
primary measure of displacement and included inequality as a supplementary
measure (fig.~\ref{fig:gini}). Across the 13 LLM populations, the mean Gini
coefficient was 0.446 in the independent condition, compared with 0.706 and
0.794 under weak and strong influence, respectively. The corresponding
increases were 0.260 and 0.348 (two-sided Wilcoxon signed-rank tests across
models, $p<0.001$ for both). The Gini coefficient exceeded the independent
baseline in all 13 models under strong influence and in 12 of 13 under weak
influence, with GPT-5 nano the only exception.

The magnitude of the increase differed substantially between reasoning classes.
Among reasoning models, the mean Gini coefficient rose from 0.528 in the
independent condition to 0.657 and 0.722 under weak and strong influence,
respectively. Among non-reasoning models, mean Gini rose from 0.395 to 0.737
and 0.838. The between-class difference in the increase under strong influence
was significant at the model level (two-sided Mann--Whitney U test,
$p<0.05$). The human benchmark showed increases in inequality of 0.153 and
0.311 under weak and strong influence, respectively.

Absolute levels of inequality require a different interpretation from KL
divergence because model populations already differed substantially from humans
and from each other before social information was introduced.
Independent-condition Gini was 0.446 across LLMs compared with 0.187 in the
human benchmark. Accordingly, the increase from each population's own
independent baseline provides a more informative comparison. Non-reasoning
models showed larger mean increases in inequality than did humans under both
influence conditions, whereas reasoning models showed smaller mean increases.

Unlike KL divergence and unpredictability, inequality did not show a
statistically significant increase across models when influence was
strengthened. Gini was higher under strong than weak influence in 9 of 13
models, but the difference did not reach conventional significance (two-sided
Wilcoxon signed-rank test, $p>0.05$). This difference between inequality and
KL divergence reflects the different information captured by the measures: two
worlds can have similar inequality while concentrating recommendations on
entirely different songs.

\subsection{Social influence attenuated but did not eliminate the correspondence between intrinsic appeal and popularity}

Table~\ref{tab:pearson_results} reports the correlation between each song's
market share under social influence and its market share under independent
evaluation. The relationship remained positive in every model and under both
influence conditions but weakened under strong influence. Across the 13 models,
the mean Pearson correlation declined from $r=0.584$ under weak influence to
$r=0.472$ under strong influence. Correlations were lower under strong
influence in 11 of 13 models, the exceptions being Claude Sonnet 4.6 and
GPT-4o. The difference was statistically significant across all models
(two-sided Wilcoxon signed-rank test, $p<0.01$).

This correspondence differed markedly across model classes. Reasoning models
retained relatively strong alignment with their independent distributions, with
mean correlations of $r=0.738$ and $r=0.689$ under weak and strong
influence, respectively, compared with 0.487 and 0.336 among non-reasoning
models. The between-class difference was significant at the model level in
both influence conditions (two-sided Mann--Whitney U tests, $n=5$ and $8$,
$p<0.05$). The human correlation under weak influence ($r=0.740$) was close
to that of the reasoning models; under strong influence, the human correlation
($r=0.557$) fell between the two class means. In the strong-influence
condition, correspondence was particularly weak in three non-reasoning models:
Gemma 3 4B ($r=0.118$), GPT-4o mini ($r=0.149$), and Qwen3-Next 80B
($r=0.151$).

In sum, exposure to cumulative recommendation counts weakened but did not
eliminate the correspondence between popularity and intrinsic appeal: songs
favored under independent evaluation generally remained more popular under
social influence.

\subsection{Early movers predicted later popularity better than intrinsic appeal}\label{app:early_movers}
 
Table~\ref{tab:pearson_results} also reports the correlation between market
shares among the first 30 agents in each influence world and market shares
among all later agents in the same world. Comparing this correlation with the
correlation between influence and independent worlds indicates whether later
popularity was better predicted by early movers or by intrinsic appeal.
 
Early-mover correlations were high in every LLM population, ranging from 0.758
to 0.979 under weak influence and from 0.792 to 0.998 under strong influence.
They exceeded the correlation with intrinsic appeal in 12 of 13 models under
weak influence, with GPT-5 the exception (0.823 vs.\ 0.885), and in all 13
models under strong influence (two-sided Wilcoxon signed-rank tests across
models, $p<0.01$ and $p<0.001$, respectively).  Mean
early-mover correlations were similar in the two classes (0.878 and 0.882 among
reasoning models and 0.866 and 0.932 among non-reasoning models under weak and
strong influence, respectively), but the gap was not. Under strong influence,
the early-mover correlation exceeded the correlation with intrinsic appeal by
0.19 on average among reasoning models and by 0.60 among non-reasoning models
(two-sided Mann--Whitney U test, $p<0.05$); under weak influence, the gaps were
0.14 and 0.38, and the class difference was not significant. The gap was
largest in Gemma 3 4B under strong influence, where the first 30 agents
predicted later market shares almost perfectly ($r=0.998$) while intrinsic
appeal predicted them weakly ($r=0.118$). In such populations, later agents
largely reproduced whatever ranking the first agents happened to establish.
 
Humans showed this pattern only under strong influence. Under weak influence,
the first 30 participants predicted later market shares less well ($r=0.316$)
than intrinsic appeal did ($r=0.740$), whereas under strong influence the
early-mover correlation (0.723) exceeded the correlation with intrinsic appeal
(0.557). This is consistent with the human self-correction reported in
fig.~\ref{fig:self_correction}: when popularity did not determine list
position, later human choices tracked intrinsic appeal more closely than the
choices of early movers. Among LLMs, early movers outweighed intrinsic appeal
under both levels of influence.

\subsection{Position bias could provide an alternative explanation for the effects of strong influence}
Language models are known to be sensitive to position in multiple-choice tasks
\citep{zheng2024robust, pezeshkpour2024order}, to use information unevenly
across the input context \citep{liu2024lost}, and to exhibit item-position bias
when ranking candidates for recommendation \citep{hou2024large}. Position bias
is also well documented in human choices from ranked lists, including
eye-tracking and click-log studies of search results
\citep{joachims2005accurately, craswell2008experimental}. In the independent
and weak-influence conditions, title order was randomized for each agent, so a
preference for early positions fell on different songs for different agents
and could not systematically favor any song. Under strong influence, however,
popularity determined position. We therefore examined the possibility that
first-stage selections in the strong-influence condition reflected a
content-independent preference for early list positions rather than a response
to popularity. 

\begin{figure}[H]
\centering
\includegraphics[width=0.9\textwidth]{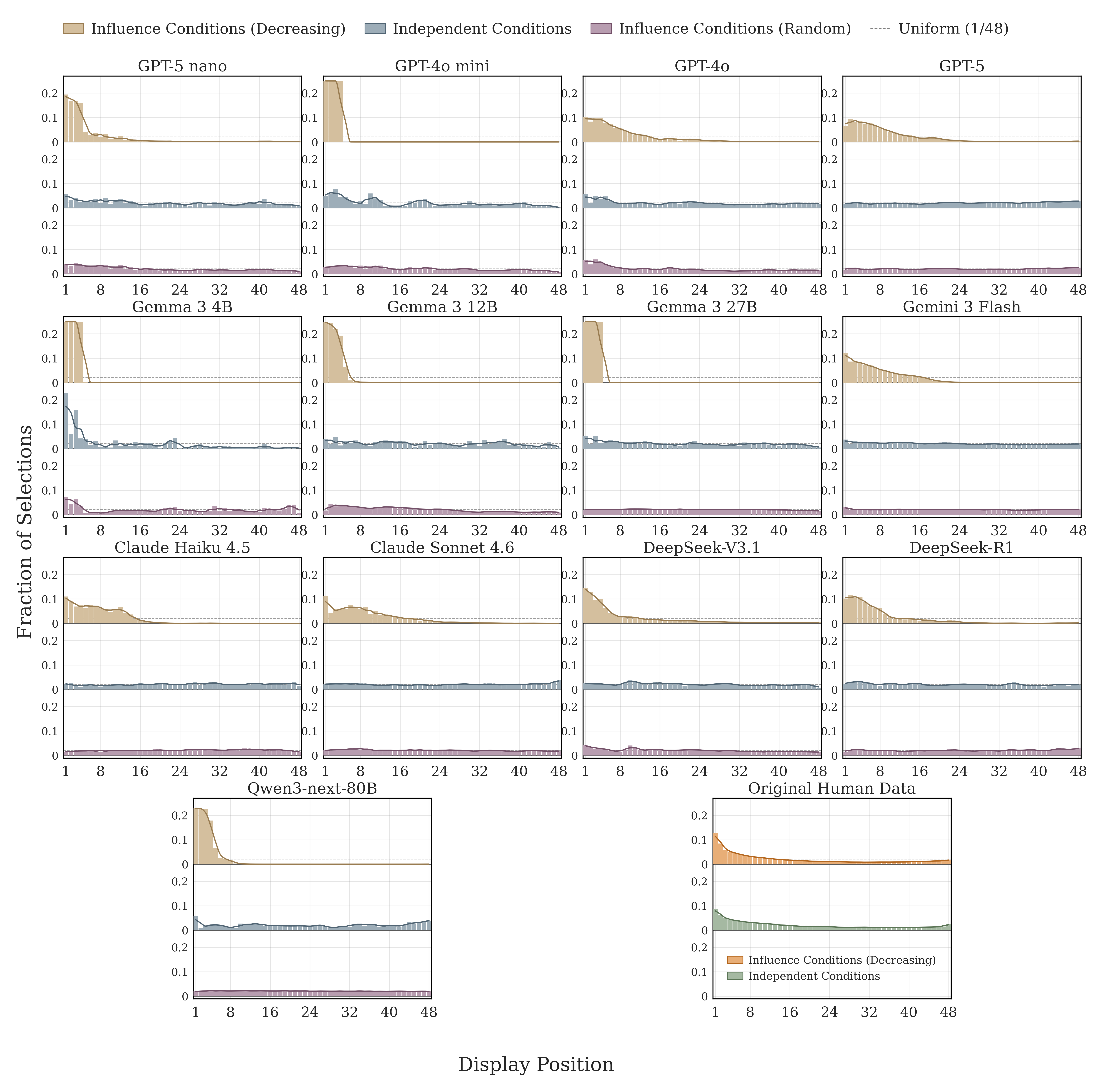}
\caption{\textbf{Position bias among LLMs and humans}. Gold, purple, and blue
denote the strong influence, weak influence, and independent conditions,
respectively. The human data show the corresponding distributions under strong
influence (orange) and independent evaluation (green). Bars show the fraction
of the first-stage selections made at each of the 48 display positions and
solid lines show the smoothed distribution. The dashed line indicates uniform
selection (equal fractions at each position). Position bias was measured in the
independent condition using the fraction of first-stage selections made at each
of the 48 display positions (not songs). For most models, the distribution
approaches uniformity in the randomly ordered independent and weak-influence
conditions, indicating a low level of position bias.  Only Gemma 3 4B and GPT-4o mini showed a modest preference for early positions, and even here, the distributions are closer to uniformity than to the strong-influence distributions.  The figure also reports stronger
position bias in the independent condition among humans than among AI agents.}
\label{fig:position_bias}
\end{figure}

Random ordering in the independent condition provided a direct diagnostic for
position bias. Fig.~\ref{fig:position_bias} shows that selections were
approximately uniform across list positions for most models, although Gemma 3
4B and GPT-4o mini showed a modest preference for early positions. The
preference for early positions was far greater among humans than among most
LLMs. (The figure does not report the positional distribution for humans in
the weak-influence condition because Music Lab used a multi-column grid, not
directly comparable to the single-column list presented to LLM agents.) A
plausible explanation is human cognitive limitations that discourage full
exploration of the entire list of 48 titles.

Because position bias is an attribute of the agents, who were randomly
allocated to all three conditions, the position bias observed in the
independent condition, which was low for most LLMs and higher for humans, may
generalize to both influence conditions. The conditions differed in whether
songs were ordered randomly or by popularity, but used the same single-column
list. This generalization could be tested directly by listing songs in
increasing order of popularity: a generic preference for early positions would
then favor the least popular songs, whereas a response to popularity would
favor songs at the bottom of the list.

If this inference is supported, part of the concentration observed under
strong influence reflected position bias acting on a popularity-ordered list
rather than a response to recommendation counts, and this component was much
larger for humans than for LLM agents. Our measures would then overstate human
susceptibility to popularity relative to that of LLMs. The weak-influence
comparison, in which position was unrelated to popularity for both
populations, is not subject to this bias. Pending an experiment that controls
for differences in position bias, the evidence in
fig.~\ref{fig:position_bias} suggests that the reported human--LLM differences
in susceptibility to cascade failure under strong influence are more likely to
be understated than overstated.

If the inference is not supported, for example because displaying
recommendation counts increased LLM agents' preference for early positions,
position bias could account for part of the strong-influence effect among
LLMs. It could not account for the weak-influence effect, however: title order
was randomized for each agent, yet every model diverged significantly from its
independent baseline and was more unpredictable than under independent
evaluation (figs.~\ref{fig:kl_divergence} and \ref{fig:unpredictability}). Nor
would it negate the importance of popularity-driven endogenous feedback. On the
contrary, path-dependent sensitivity to initial recommendations means that
even a small position bias could tip a multi-agent system to converge
arbitrarily on whatever appeared at the top of the list early on.

\subsection{Trajectories of cumulative advantage}
 
\begin{figure}[H]
\centering
\includegraphics[width=\textwidth]{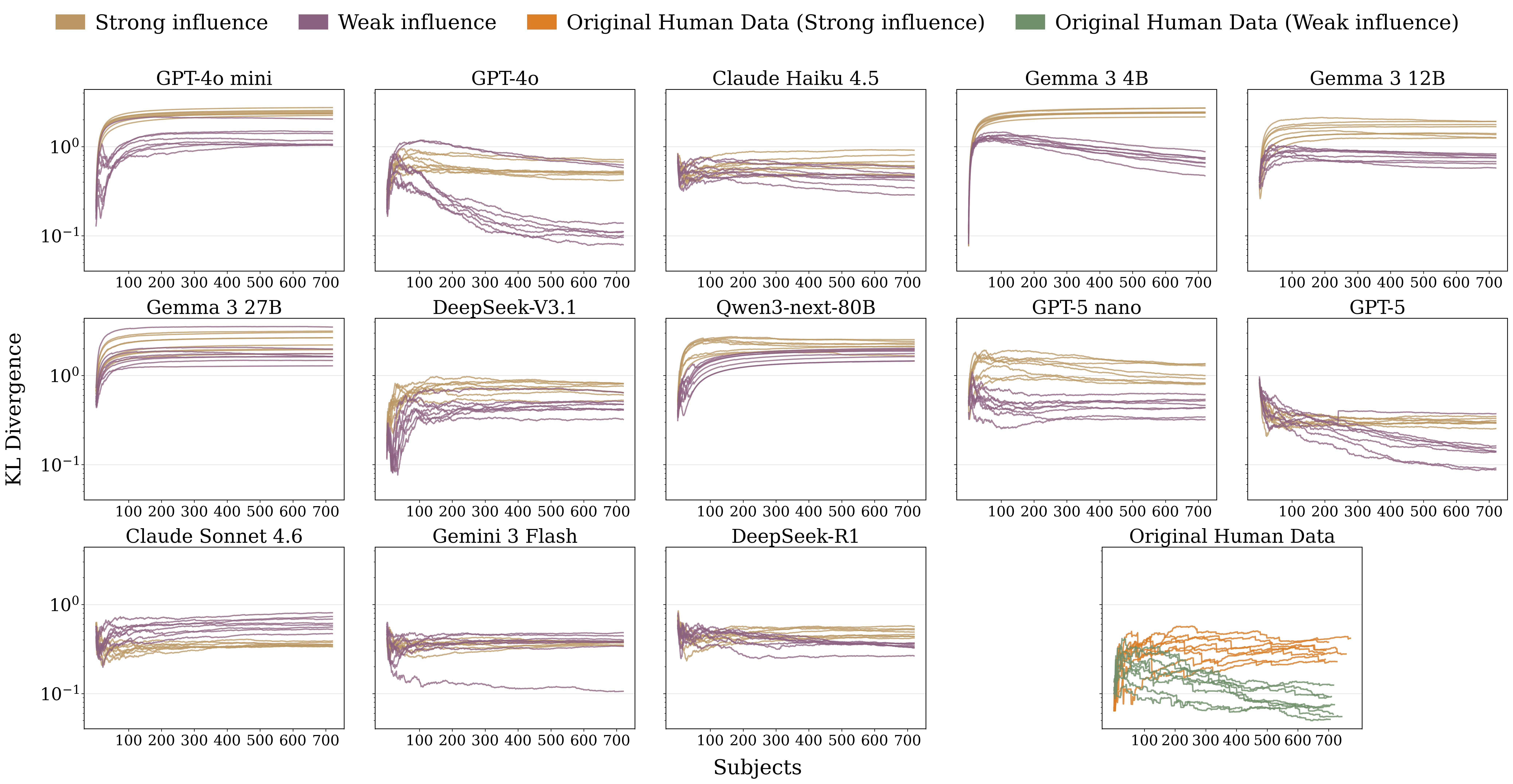}
\caption{\textbf{Dynamics of KL divergence as recommendations accumulated.} The plots display KL divergence from the corresponding independent
distribution over the course of the experiment for the 13 LLM populations and
the human benchmark, under weak and strong influence. Each line shows one of
the eight worlds in an influence condition. The horizontal axis is the number
of participants (humans or agents) who had completed the task; the vertical
axis is logarithmic. Market shares at each stage were calculated from
accumulated recommendations for LLMs and downloads for humans. Gold and purple
denote strong and weak influence for LLMs; orange and green denote strong and
weak influence for humans. Final divergence values correspond to those
reported in fig.~\ref{fig:kl_divergence}, where they are averaged over the
eight worlds in each influence condition. Divergence persisted over the
observed trajectories, although its development varied across models and
worlds.}
\label{fig:kl_dynamics}
\end{figure}

\begin{figure}[H]
\centering
\includegraphics[width=\textwidth]{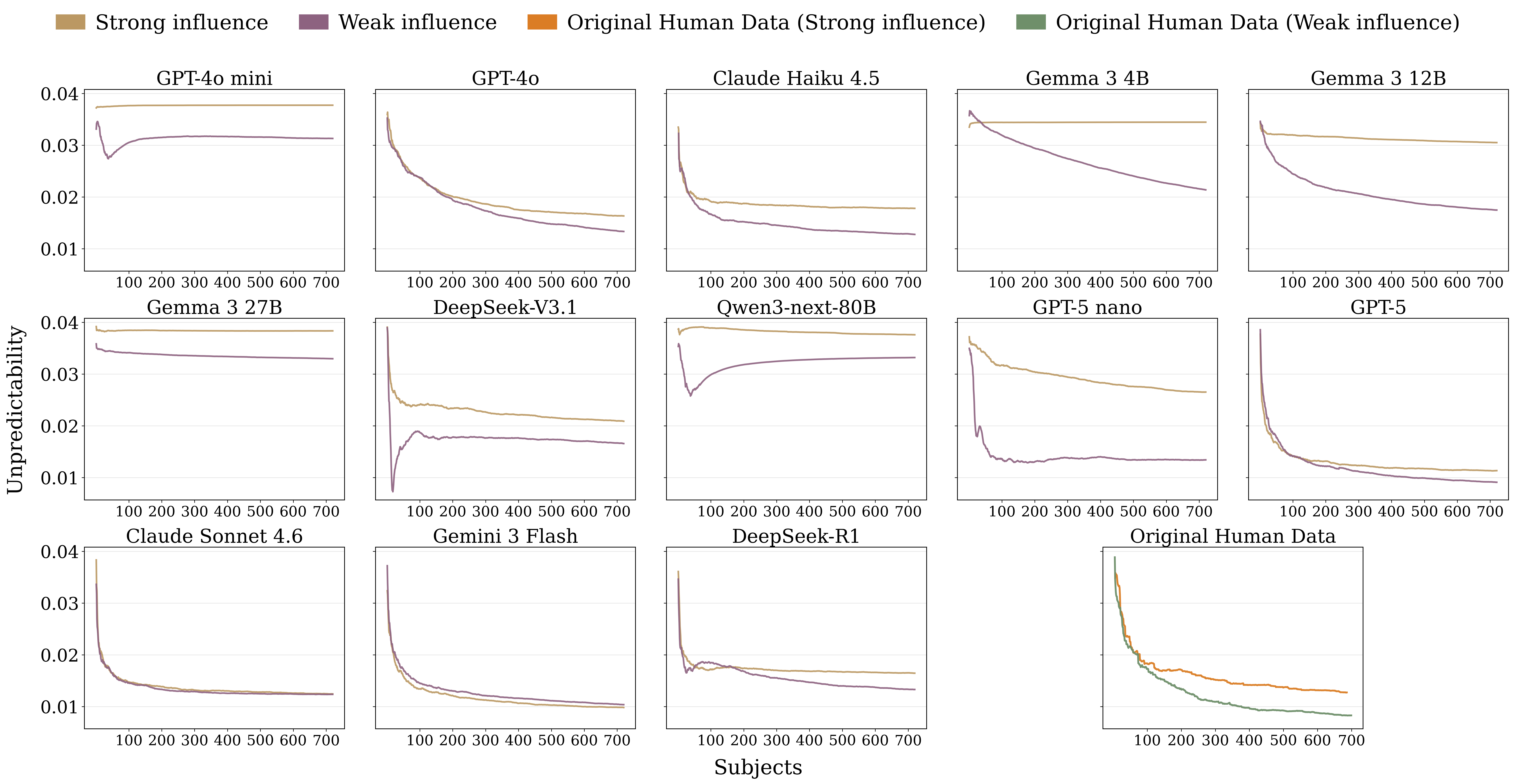}
\caption{\textbf{Dynamics of unpredictability as recommendations accumulated.} The plots report unpredictability over the course of the experiment for the
13 LLM populations and the human benchmark, under weak and strong influence.
Unpredictability measures the mean absolute difference in a song's market
share across all 28 pairs of the eight worlds in an influence condition,
averaged over the 48 songs, with market shares calculated from the
recommendations (downloads for humans) accumulated at that stage. The
horizontal axis is the number of participants (agents or humans) who had
completed the task. Gold and purple denote strong and weak influence for
LLMs; orange and green denote strong and weak influence for humans. Final
unpredictability values correspond to those reported in
fig.~\ref{fig:unpredictability}. Unpredictability generally stabilized or
declined over time, with strong influence producing greater final
unpredictability than weak influence in most models.}
\label{fig:u_dynamics}
\end{figure}

We examined the dynamics of KL divergence and unpredictability as each world
evolved (figs.~\ref{fig:kl_dynamics} and \ref{fig:u_dynamics}). These
trajectories show how divergence from independent judgment and differences
among influence worlds developed over the course of the experiment, using
market shares calculated from the choices accumulated at each stage. Final
values correspond to the outcomes reported in figs.~\ref{fig:kl_divergence}
and \ref{fig:unpredictability}. Early values were based on fewer choices and
were consequently more sensitive to individual recommendations.
 
\subsection{Robustness analyses}

Two features of the experimental design could affect the magnitude of social
influence: sampling temperature and the number of alternative songs each agent
was allowed to consider for possible recommendation. We therefore repeated the
multiple-world experiment under alternative settings for each parameter. Both
robustness checks used the independent and strong-influence conditions.

\paragraph{Robustness to sampling temperature.}

\begin{figure}[H]
\centering
\includegraphics[width=\textwidth]{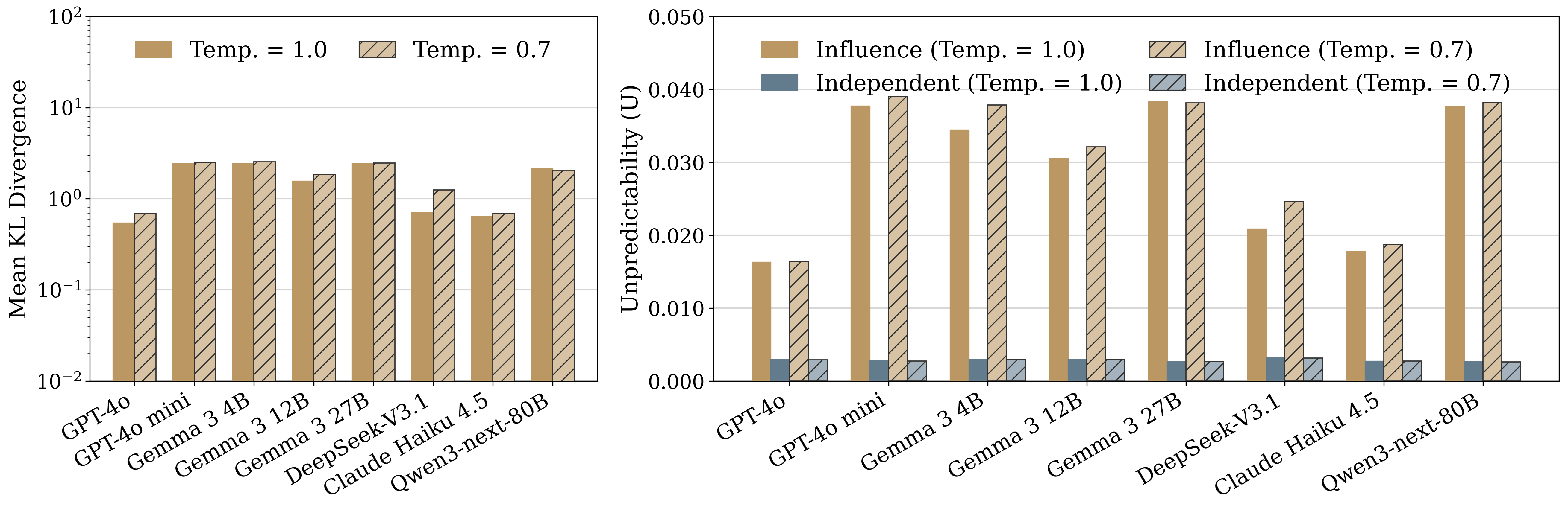}
\caption{\textbf{Effects of social influence were robust to a lower sampling
temperature.} The figure reports KL divergence (left) and unpredictability
(right) for the eight non-reasoning models under strong influence at
temperatures 1.0 (solid bars) and 0.7 (hatched bars). KL divergence compares
each influence world with the independent baseline generated at the same
temperature, averaged over the eight influence worlds; the vertical axis is
logarithmic. Unpredictability is shown for both influence (gold) and
independent (blue) conditions; influence values are the mean absolute
difference in a song's market share across all 28 pairs of the eight influence
worlds, averaged over the 48 songs. Outcomes in the independent condition were
estimated from 1,000 random splits into two equal subpopulations. Lowering
temperature did not attenuate the effects of social influence on average, and
unpredictability in the independent condition hardly changed. Averaged over the
eight models, KL divergence was 1.62 at temperature 1.0 and 1.75 at temperature
0.7, with greater divergence at the lower temperature in seven of the eight
models. Unpredictability in the influence condition averaged 0.0292 and 0.0306,
respectively, compared with 0.00288 and 0.00284 in the independent condition.}
\label{fig:temperature_robustness}
\end{figure}

The main experiment used a sampling temperature of 1.0 for non-reasoning
models. Because temperature controls sampling variability, the observed
divergence and path dependence could depend on this inference setting. We
therefore repeated the experiment at temperature 0.7 for all eight
non-reasoning models (fig.~\ref{fig:temperature_robustness}). We restricted
this manipulation to non-reasoning models because the reasoning models used
model-specific reasoning controls, making temperature an inconsistent
manipulation.

Lowering the temperature did not attenuate overall susceptibility to social
influence. Mean KL divergence across the eight models was 1.62 at temperature
1.0 and 1.75 at temperature 0.7, increasing at the lower temperature in seven
of eight models. Model ordering by KL divergence was identical across
temperatures (Spearman $\rho=1.00$). Influence-world unpredictability also
remained high, increasing from $U=0.0292$ to 0.0306. Independent-condition
unpredictability was nearly unchanged (0.00288 versus 0.00284). 

\paragraph{Robustness to the size of the consideration set.}

\begin{figure}[H]
\centering
\includegraphics[width=\textwidth]{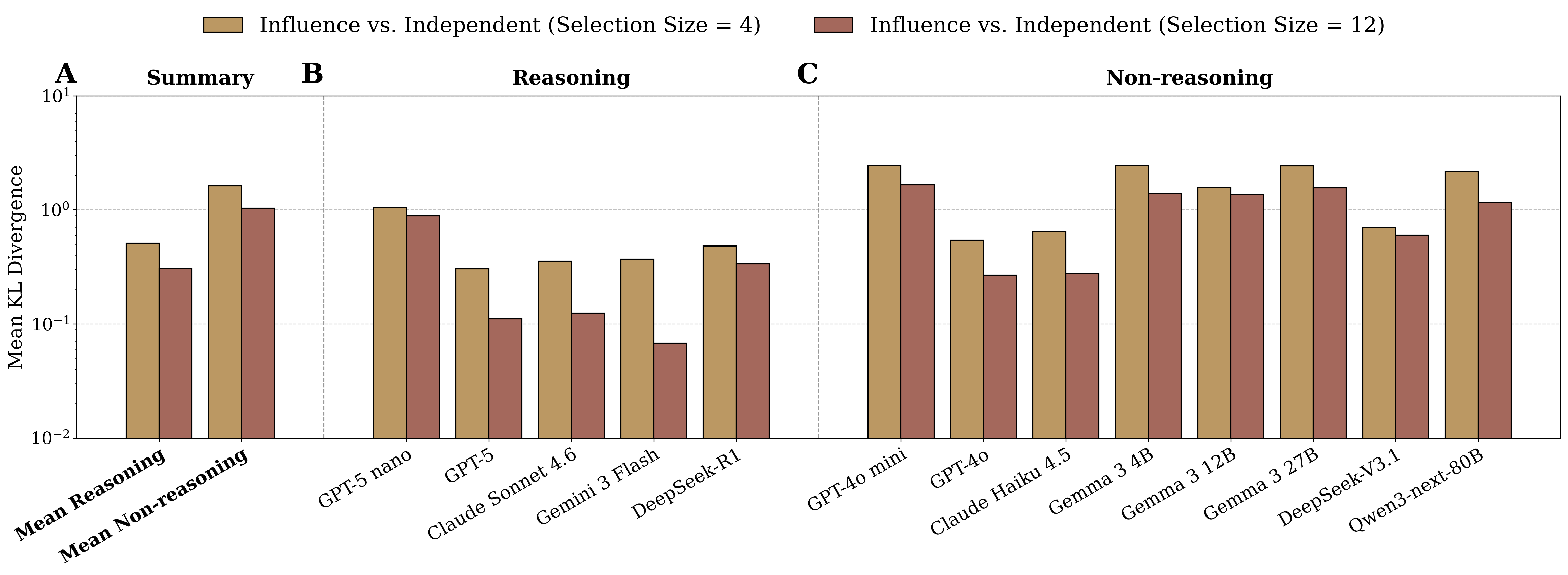}
\caption{\textbf{KL divergence was robust to an increase in the number of
songs that could be selected for assessment.} The figure reports KL divergence
between strong-influence worlds and their matched independent baselines when
agents selected four (gold) or 12 (brown) of the 48 songs for evaluation. Each
model's divergence is averaged over eight influence worlds. In panel A, bars
indicate the unweighted means across the five reasoning and eight
non-reasoning models. The vertical axis is logarithmic. Although the larger
selection budget reduced divergence in all 13 models, the ratio of mean divergence in non-reasoning to reasoning models was nearly unchanged, as shown by the similar vertical distances between the two gold bars and between the two brown bars in panel A.}
\label{fig:selection_kl_robustness}
\end{figure}

\begin{figure}[H]
\centering
\includegraphics[width=\textwidth]{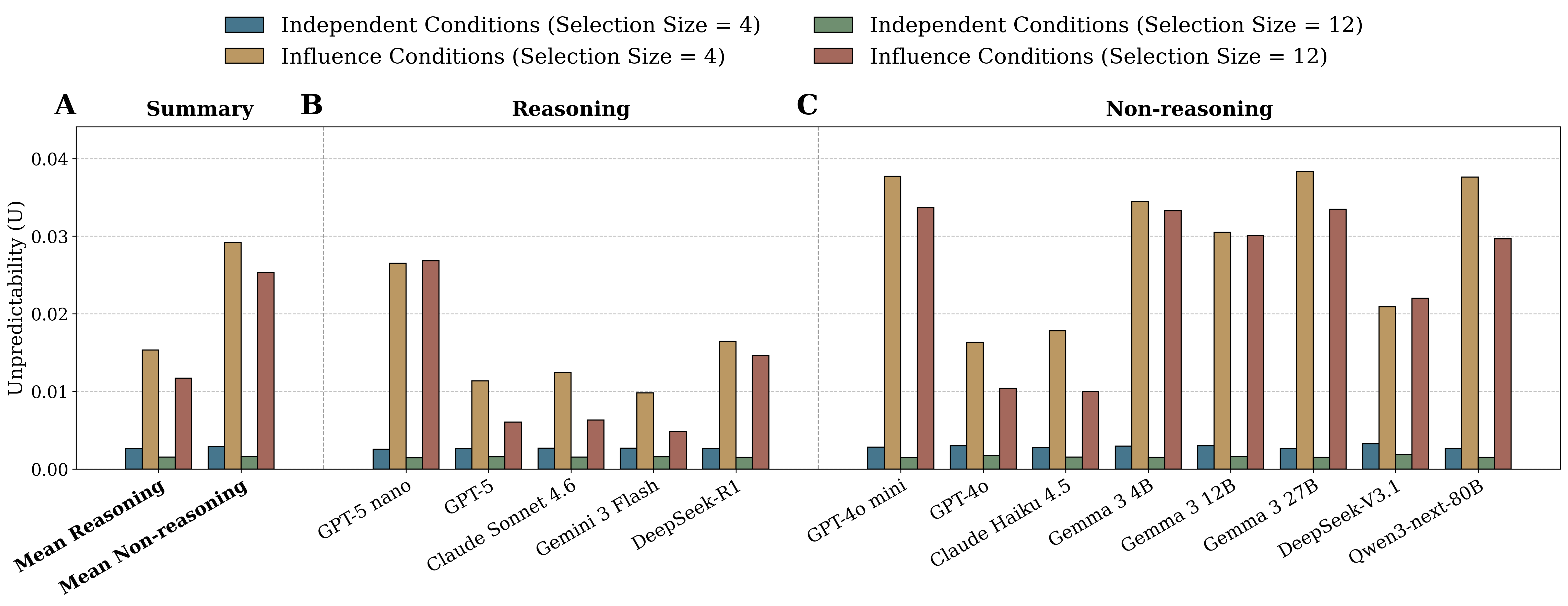}
\caption{\textbf{Unpredictability was robust to an increase in the number of
songs that could be selected for assessment.} The figure reports
unpredictability under strong influence and independent evaluation when the
number of songs agents could evaluate was increased from four to 12 out of 48.
Unpredictability in the influence condition is the average absolute difference
in market shares for the 48 songs across all 28 pairs of eight worlds.
Unpredictability in the independent condition is the average of 1,000 random
splits into two groups of 720 agents. \textbf{A.} Bars show unweighted means across the
five reasoning and eight non-reasoning models. Gold and brown denote strong
influence with four and 12 selections, respectively; blue and green denote the
corresponding independent conditions. Increasing the selection budget reduced
unpredictability on average across the 13 models (panel A), but every
influence population remained more unpredictable than every independent
population at both selection sizes.}
\label{fig:selection_u_robustness}
\end{figure}

Agents in the primary experiment selected four of the 48 songs for evaluation,
approximating the average number assessed by participants in the original Music
Lab experiment \citep{salganik2006experimental}. To test whether the results
depended on the size of the assessment set, we repeated the experiment for all
13 models, changing only the number of songs that could be selected for
assessment from four to 12 (figs.~\ref{fig:selection_kl_robustness} and
\ref{fig:selection_u_robustness}).

Increasing the assessment set attenuated but did not eliminate divergence. Mean
KL divergence decreased from 1.19 with four selections to 0.75 with 12, with a
reduction in all 13 models. Model ordering was strongly preserved across
selection sizes (Spearman $\rho=0.95$), as was the difference between model
classes: with 12 selections, mean KL divergence was 0.31 among reasoning models
and 1.03 among non-reasoning models.

Influence-world unpredictability also decreased, but only slightly, from
$U=0.0239$ to $U=0.0201$, with reductions in 11 of 13 models and strongly
preserved model ordering (Spearman $\rho=0.96$). Independent-condition
unpredictability also decreased, from 0.00278 to 0.00156. With 12 selections, the lowest value
observed in an influence condition ($U=0.0048$) was approximately 2.6 times
the highest value observed in the independent condition ($U=0.0019$).
Relative to each model's corresponding independent baseline, mean
$U_{\mathrm{inf}}/U_{\mathrm{ind}}$ was 8.6 with four selections and 13.0
with 12 selections.

\clearpage
\subsection{Limitations and directions for future work}

Several questions remain open. First, agents observed accumulated
recommendation counts without discussing choices or retaining a shared
conversational history. Future work could examine whether agents that exchange
arguments, remember earlier discussions, reconsider judgments, or incentivize
conformity amplify or attenuate the dynamics of cumulative advantage.

Second, each population consisted of a single model, with heterogeneity
introduced through personas. Future work could examine further increases in
agent heterogeneity, by mixing together models of differing capability, and by
including both humans and AI agents in each world.

AI-assisted scientific peer review is another important application, where
agents recommend academic papers instead of songs. Future experiments could
test whether exposure to earlier recommendations improves the quality of
evaluations or produces self-reinforcing cascades of acceptance and rejection.

\end{document}